\documentclass[11pt,letterpaper]{article}

\usepackage{xcolor}
\definecolor{navy}{rgb}{0,0.1,0.4}
\definecolor{navy-cap}{rgb}{0,0.1,0.5}
\definecolor{navy-ref}{rgb}{0,0.3,1}
\definecolor{lgray}{gray}{0.90}
\definecolor{dkgreen}{rgb}{0,0.6,0}
\definecolor{gray}{rgb}{0.5,0.5,0.5}
\definecolor{mauve}{rgb}{0.58,0,0.82}

\usepackage{sectsty}
\sectionfont{\fontsize{12}{15}\selectfont \color{navy}}
\subsectionfont{\fontsize{12}{15}\selectfont \color{navy}}
\subsubsectionfont{\fontsize{12}{15}\selectfont \color{navy}}

\usepackage[margin = 1 in]{geometry}

\usepackage{times}
\usepackage[font={color=navy-cap, small}, labelfont={bf}, labelsep=space, justification=justified, skip=0pt]{caption}
\usepackage[bottom, hang,flushmargin]{footmisc} 

\usepackage{titlesec}
\usepackage{tikz}
\titlespacing{\section}{0pt}{0.75em}{0.25em}
\titlespacing{\subsection}{0pt}{0.5em}{0.25em}
\titlespacing{\subsubsection}{0pt}{0.5em}{0.25em}

\usepackage{array,longtable}
\usepackage{xltabular}
\usepackage{multirow}
\usepackage{pifont}
\usepackage[font={color=navy-cap, footnotesize}, labelfont={bf}, labelsep=space, justification=justified, skip=3pt]{caption}
\usepackage{subcaption}
\usepackage{booktabs} 
\usepackage{graphicx}
\usepackage{wrapfig}
\usepackage{enumitem}
\usepackage{hyperref}
\hypersetup{colorlinks = true, linkcolor = navy-ref}
\usepackage{textcomp}
\usepackage{url}
\usepackage{amssymb}
\usepackage{bm}
\usepackage{amsmath}

\usepackage{dirtytalk}
\usepackage{pdfpages} 
\usepackage{soul} 
\usepackage{fancyhdr}
\usetikzlibrary{patterns}
\usepackage{pgfplots}
\usepackage{floatrow}
\usepackage{setspace}
\usepackage{bbm}

\usepackage[
backend=biber,
style=numeric-comp,
sorting=none,
maxnames = 10
]{biblatex}
\usepackage[nameinlink]{cleveref} 

\usepackage{colortbl}
\usepackage{hhline}

\newcommand{\cmt}[1]{} 
\newcommand\brackets[1]{\mathopen{}\left[#1\right]\mathclose{}}
\newcommand\parens[1]{\mathopen{}\left(#1\right)\mathclose{}}
\newcommand\bars[1]{\left|#1\right|}
\newcommand\braces[1]{\mathopen{}\left\{#1\right\}\mathclose{}}

\newcommand{\Ab}{\boldsymbol{A}}

\newcommand{\Cb}{\boldsymbol{C}}

\newcommand{\CYb}{\boldsymbol{C}_y}
\newcommand{\Cphib}{\boldsymbol{C}_\phi}

\DeclareMathOperator*{\E}{\mathbb{E}}
\newcommand{\eg}{e.g., }

\newcommand{\ie}{i.e., }

\newcommand{\mb}{\boldsymbol{m}}

\newcommand{\Mb}{\boldsymbol{M}}

\newcommand{\mathC}{\mathcal{C}}
\newcommand{\mathD}{\mathcal{D}}

\newcommand{\mathGdagger}{\mathcal{G}^\dagger}
\newcommand{\mathGtildedagger}{\widetilde{\mathcal{G}}^\dagger}
\newcommand{\mathGtildedaggerp}{\widetilde{\mathcal{G}}^\dagger_p}

\newcommand{\mathL}{\mathcal{L}}

\newcommand{\mathO}{\mathcal{O}}
\newcommand{\mathT}{\mathcal{T}}
\newcommand{\mathU}{\mathcal{U}}
\newcommand{\mathUstar}{\mathcal{U}^*}
\newcommand{\mathV}{\mathcal{V}}
\newcommand{\mathVstar}{\mathcal{V}^*}

\newcommand{\real}{\mathbb{R}}

\newcommand{\ut}{u_t}
\newcommand{\ux}{u_x}
\newcommand{\uy}{u_y}
\newcommand{\uxx}{u_{xx}}
\newcommand{\uyy}{u_{yy}}

\newcommand{\vx}{v_{x}}
\newcommand{\vy}{v_{y}}
\newcommand{\vxx}{v_{xx}}
\newcommand{\vyy}{v_{yy}}

\newcommand{\px}{p_{x}}
\newcommand{\py}{p_{y}}

\newcommand{\Ub}{\boldsymbol{U}}

\newcommand{\Vb}{\boldsymbol{V}}

\newcommand{\vb}{\boldsymbol{v}}

\newcommand{\vecOp}{\text{vec}}

\newcommand{\Xb}{\boldsymbol{X}}

\newcommand{\Yb}{\boldsymbol{Y}}
\newcommand{\Ystarb}{\boldsymbol{Y^*}}

    \makeatletter
\def\@fnsymbol#1{\ensuremath{\ifcase#1\or \dagger\or *\or \ddagger\or
   \mathsection\or \mathparagraph\or \|\or **\or \dagger\dagger
   \or \ddagger\ddagger \else\@ctrerr\fi}}
    \makeatother

\newcommand{\nosymbolfootnote}[1]{%
  \renewcommand{\thefootnote}{}%
  \footnote{#1}%
  \renewcommand{\thefootnote}{\arabic{footnote}}
}

\title{Operator Learning with Gaussian Processes}
\date{\vspace{-5ex}}
\usepackage{authblk}
\author[1]{Carlos Mora}
\author[1]{Amin Yousefpour}
\author[1]{Shirin Hosseinmardi}
\author[2]{Houman Owhadi}
\author[1]{Ramin Bostanabad\thanks{\noindent Corresponding Author: Raminb@uci.edu}}
\affil[1]{Department of Mechanical and Aerospace Engineering, University of California, Irvine, CA, United States of America}
\affil[2]{Computing and Mathematical Sciences, Caltech, Pasadena, CA, United States of America}

\begin{document}
    \pagenumbering{arabic}
    \maketitle
    \sloppy 
    \noindent \textcolor{navy}{\textbf{Abstract}}

Operator learning focuses on approximating mappings $\mathcal{G}^\dagger: \mathcal{U} \rightarrow \mathcal{V}$ between infinite-dimensional spaces of functions, such as $ u: \Omega_u \rightarrow \mathbb{R} $ and $v: \Omega_v \rightarrow \mathbb{R}$. This makes it particularly suitable for solving parametric nonlinear partial differential equations (PDEs). Recent advancements in machine learning (ML) have brought operator learning to the forefront of research. While most progress in this area has been driven by variants of deep neural networks (NNs), recent studies have demonstrated that Gaussian Process (GP)/kernel-based methods can also be competitive. These methods offer advantages in terms of interpretability and provide theoretical and computational guarantees.
In this article, we introduce  a hybrid GP/NN-based framework for operator learning, leveraging the strengths of both deep neural networks and kernel methods.
 Instead of directly approximating the function-valued operator $\mathcal{G}^\dagger$, we use a GP to approximate its associated real-valued bilinear form 
$
\mathGtildedagger: \mathcal{U} \times \mathcal{V}^* \rightarrow \mathbb{R}.
$
This bilinear form is defined by the dual pairing 
$
\mathGtildedagger(u, \varphi) := [\varphi, \mathGdagger(u)],
$
which allows us to recover the operator \( \mathGdagger \) through 
$
\mathGdagger(u)(y) = \mathGtildedagger(u, \delta_y).
$
The mean function of the GP can be set to zero or parameterized by a neural operator and for each setting we develop a robust and scalable training mechanism based on maximum likelihood estimation (MLE) that can optionally leverage the physics involved. 
Numerical benchmarks demonstrate its scope, scalability, efficiency, and robustness, showing that (1) it enhances the performance of a base neural operator by using it as the mean function of a GP, and (2) it enables the construction of zero-shot data-driven models that can make accurate predictions without any prior training. 
Additionally, our framework (a) naturally extends to cases where $\mathGdagger: \mathcal{U} \rightarrow \prod_{s=1}^S \mathV^s$ maps into a vector of functions, and (b) benefits from computational speed-ups achieved through product kernel structures and Kronecker product matrix representations of the underlying kernel matrices\nosymbolfootnote{\hspace{1.75em}GitHub repository: \href{https://github.com/Bostanabad-Research-Group/GP-for-Operator-Learning}{https://github.com/Bostanabad-Research-Group/GP-for-Operator-Learning}}.

\noindent \textcolor{navy}{\textbf{Keywords:}} 
Operator Learning; Gaussian Processes; Neural Operators; Zero-shot Learning; Optimal Recovery. 

    \section{Introduction} \label{sec intro}
Operator learning naturally arises in many applications such as solving partial differential equation (PDE) systems \cite{ghanem2003stochastic}, statistical description and modeling of random functions \cite{ramsay2005fitting, ramsay2002applied}, mechanistic reduced order modeling \cite{lucia2004reduced}, speech
inversion and sound recognition \cite{mitra2009acoustics}, or emulation of expensive simulations \cite{economon2016su2, kovachki2022multiscale, boncoraglio2021active}. Recent advancements in machine learning (ML) have pushed operator learning to the forefront of  research in both academia and industry. While the vast majority of the developments on this topic leverage variants of deep neural networks (NNs) \cite{raonic2023convolutional, lu2021learning, li2020fourier, hao2023gnot},  kernel-based methods \cite{batlle2024kernel} have also recently been shown to be competitive.
In this paper, we propose a hybrid GP/NN-based framework for approximating mappings between infinite-dimensional function spaces.
Our framework provides a first of its kind mechanism for simultaneously leveraging the strengths of both deep NNs and kernels for operator learning.

\subsection{Description of the Operator Learning Problem} \label{subsec problem description}
Let $\mathU$ and $\mathV$ denote two separable infinite-dimensional Banach spaces of continuous functions and define the nonlinear operator $\mathGdagger$ as:
\begin{equation}
    \mathGdagger: \mathU \rightarrow \mathV.
    \label{eq operator learning problem}
\end{equation}
Furthermore, let $\phi:\mathU \rightarrow \real^p$ and $\psi:\mathV \rightarrow \real^q$ be two bounded linear observation operators, that is:
\begin{equation}
    \phi: u \mapsto u^p := \left [ u(x_1), \ldots, u(x_p)\right]^T 
    \quad \text { and } \quad 
    \psi: v \mapsto v^q := \left[v(y_1), \ldots, v(y_q)\right]^T
    \label{eq observation operators}
\end{equation}
where $\{x_j\}_{j=1}^p$ and $\{y_j\}_{j=1}^q$ denote the location of the points where the functions $u$ and $v$ are observed in their respective domains $\Omega_u$ and $\Omega_v$.
Our goal is to approximate $\mathGdagger$ using $N$ pairs of input-output observations, i.e., we aim to approximate $\mathGdagger$ given $\mathD := \{\phi(u_i), \psi(v_i)\}_{i=1}^N$ where $\{u_i,v_i \}_{i=1}^N$ are elements from $\mathU \times \mathV$.

\subsection{Summary of the Proposed Approach} \label{subsec method summary}
We propose to cast the operator learning problem in \Cref{eq operator learning problem} as the following equivalent functional learning problem:
\begin{equation}
    \mathGtildedagger: \mathU \times \mathVstar \rightarrow \real
    \label{eq functional learning problem}
\end{equation}
where $\mathVstar$ is the dual space of $\mathV$, consisting of all continuous linear functionals from $\mathV$ to $\real$. 
For any pair $(u,\varphi)\in \mathU \times \mathVstar$, $\mathGtildedagger(u,\varphi)$ returns the scalar value $[\varphi,\mathGdagger(u)]$ defined as the dual pairing between $\varphi$ and $\mathGdagger(u)$. If we specifically choose $\varphi=\delta_y$, where $\delta_y$ is the delta functional which evaluates the output function at a specific point $y\in \Omega_v \subset \real^d$, then we can easily retrieve the desired operator $\mathGdagger$ given that $\mathGdagger(u)(y) = \mathGtildedagger(u,\delta_y)$. 

In practice, (1) we only have access to a discretization of $u$, \ie $\phi(u)\in \real^p$, (2) we are only interested in pointwise evaluations  $\mathGdagger(u)(y)$ with $y\in \Omega_v$, 
and hence the functional learning problem can be reduced to identifying an operator:
\begin{equation}
    \mathGtildedaggerp: \real^p \times \Omega_v \rightarrow \real.
    \label{eq function learning problem}
\end{equation}
such that  $\mathGtildedaggerp\big(\phi(u),y\big)\approx \mathGdagger(u)(y)$.

\begin{figure}[!t]
    \centering
    \includegraphics[width=0.9\linewidth]{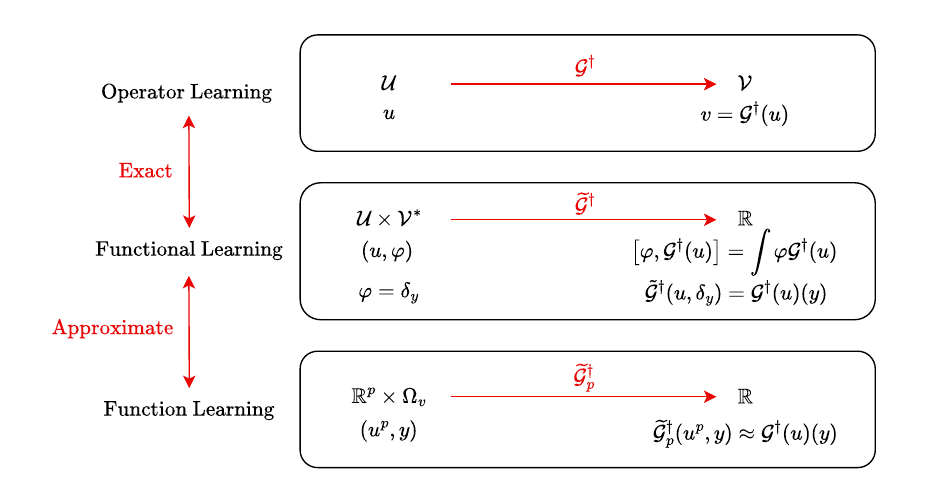}
    \caption{\textbf{Diagram of our framework for operator learning:} We convert the operator learning problem to a regression one which can be solved via GPs. For multi-output operator learning, we use multi-response or multi-task GPs.}
    \label{fig diagram}
\end{figure}

Since the mapping in \Cref{eq function learning problem} is between finite-dimensional Euclidean spaces it can be numerically approximated via classical NN/GP-based regression techniques. This regression-based approach differs from other kernel-based methods such as \cite{batlle2024kernel} that directly approximate the operator rather than its dual action on $\mathcal{V}^*$. The overall articulation of our regression-based approach for operator learning is schematically shown in \Cref{fig diagram} which also applies to other well-known techniques such as DeepONet \cite{lu2021learning}. In this paper, we approximate the mapping $\mathGtildedaggerp$ in \Cref{eq function learning problem} via GPs whose parameters are estimated either in a purely data-driven manner, or via both data and physics. In the former case, we use maximum likelihood estimation (MLE) for parameter optimization and set the mean function to be either zero or a deep NN. 
In the latter case, we rely on GPs whose mean functions are represented via neural operators (e.g., FNO or DeepONet) whose parameters are estimated via a weighted combination of MLE and mean squared error (MSE).

As in \cite{batlle2024kernel},  the diagram in \Cref{fig diagram} can be made commutative by introducing two reconstruction (optimal recovery) operators associated with $\phi$ and $\psi$ that map the observations in $\real^p$ and $\real^q$ back to $\mathU$ and $\mathV$, respectively. Since $\phi$ and $\psi$ are bounded and linear, their elements are in $\mathUstar$ and $\mathVstar$ which are the dual spaces of $\mathU$ and $\mathV$, respectively. Assuming $\mathU$ and $\mathV$ are reproducing kernel Hilbert spaces (RKHSs) and the elements of the observation operators are linearly independent, the reconstruction (optimal recovery) operators can be obtained via the usual representer theorem as detailed in \cite{batlle2024kernel}.

\subsection{Illustrative Example} \label{subsec intro example}
To contextualize the above abstract definitions, we consider the Burgers' equation with Dirichlet boundary conditions (BCs):
\begin{equation}
    \begin{aligned}
        &v_t + vv_x - \nu v_{xx} = 0, && \forall x \in (-1,1), &&t \in (0, 1] \\
        &v = 0, && \forall x = 0, &&t \in [0, 1] \\
        &v = 1, && \forall x = 1, &&t \in [0, 1] \\
        &v = u, && \forall x \in [-1,1], &&t = 0,
    \end{aligned}
    \label{eq burgers dbc}
\end{equation}
where $x$ and $t$ denote space and time, $v$ is the PDE solution, $u$ denotes the initial condition (IC), and $\nu=0.1$ is the kinematic viscosity. We denote the $(0, 1)^2$ domain and its boundaries where IC and BCs are specified by $\Omega$, $\partial \Omega_t$, and $\partial \Omega_x$, respectively. 
The goal in this problem is to learn the nonlinear operator $\mathGdagger$ that maps $u$ to $v$, i.e., to the solution field in $\Omega$. To approximate $\mathGdagger$, we consider two scenarios:
\begin{itemize}
    \item \textbf{Data-driven operator learning:} We take the linear operators $\phi$ and $\psi$ defined in \Cref{subsec problem description} to be the function evaluations at a set of $p$ and $q$ collocation points (CPs) on $\partial \Omega_t$ and in $\Omega$, respectively. This choice provides the training dataset $\mathD := \{\phi(u_i), \psi(v_i)\}_{i=1}^N$ where $\phi(u_i)$ and $\psi(v_i)$ are of size $p$ and $q$, respectively, and the pair $\{u_i, v_i\}$ satisfies the PDE system in \Cref{eq burgers dbc}. In this scenario, we follow the diagram in \Cref{fig diagram} and approximate $\mathGdagger$ via $\mathGtildedaggerp$ using $\mathD$.
    
    \item \textbf{Physics-informed operator learning:} In this scenario, we leverage the PDE system in \Cref{eq burgers dbc} as well as $\mathD$ to approximate $\mathGdagger$. That is, we require the $N_{pi}$ samples $\mathD_{pi} := \{u_i, \widehat{v}_i\}_{i=1}^{N_{pi}}$, where the elements of $\widehat{v}_i$ are predicted by $\mathGtildedaggerp$, to satisfy the PDE system in \Cref{eq burgers dbc}.
\end{itemize}

\begin{figure}[!h]
    \centering
    \includegraphics[width=1.0\linewidth]{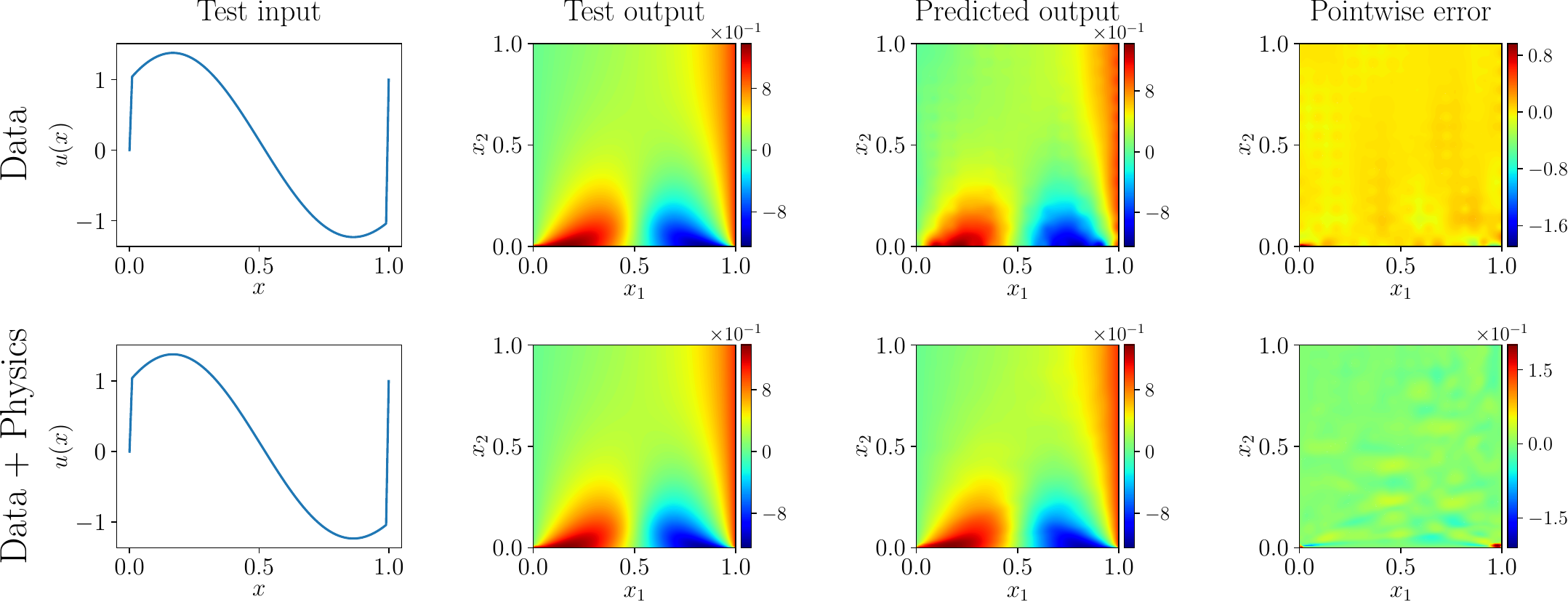}
    \caption{\textbf{Data-driven and physics-informed operator learning for the Burgers' problem with Dirichlet BCs:} The training dataset has $400$ pairs of $\{ u_i, v_i \}$ that satisfy \Cref{eq burgers dbc}. The observation operators $\phi$ and $\psi$ sample $u_i$ and $v_i$ at $p=100$ and $q=12^2$ collocation points, respectively. It can be observed that leveraging the physics reduces the prediction error. A DeepONet is used as the mean function of the GP in both data-driven and physics-informed cases.}
    \label{fig illustrative burgers dbc}
\end{figure}

We set $N = 400, p = 100, q=12^2, N_{pi} = 50$, and consider $\phi$ and $\psi$ to be pointwise function evaluations. The GPs' training mechanisms and the experimental setup for both scenarios are detailed in \Cref{sec method,sec results}, respectively. We provide example train and test samples along with our predictions in \Cref{fig illustrative burgers dbc}. It is observed that $(1)$ the predictions achieve small relative $L2$ errors which are $7.35\%$ and $1.65\%$ in the data and physics-informed cases, respectively, and $(2)$ leveraging the physics increases the approximation accuracy. 

\subsection{Review of Related Literature} \label{subsec related works}
Operator learning aims at approximating mappings between infinite-dimensional function spaces and is particularly well suited for solving parametric nonlinear PDEs. Classical operator learning approaches that are based on polynomial chaos expansion \cite{RN2055}, homogenization theories \cite{RN2056,RN2058}, or the stochastic finite element method (FEM) \cite{RN2057,RN499,RN563} have been extensively studied over the past few decades. In a broad sense, these methods represent both the input function and the PDE solution with a set of suitable bases and, in turn, cast the operator learning problem as a high-dimensional regression one where the coefficients of the input bases are mapped to those of the output bases.

\noindent \textbf{Neural Operators:} 
Fueled by their success in many scientific applications and the advancements in hardware and software, deep NNs are increasingly used for operator approximation. One of the main distinguishing factors among different methods is the network architecture which may be designed via multi-layer perceptrons, convolution layers, attention mechanism, skip connections, graphs, or many other alternatives \cite{raonic2023convolutional, lu2021learning, hao2023gnot, li2020neural}. We review some of the most prominent methods below and refer the interested reader to \cite{kovachki2024operator, boulle2023mathematical, kovachki2023neural} for more information. 

Following the universal approximation theorem for operators \cite{RN1421}, a standard DeepONet \cite{RN1380} approximates an operator via a deep NN that combines two sub-networks that encode the input function (branch net) and the spatiotemporal coordinates (trunk net). Since their inception, DeepONets have been extended in a few directions and a particular one that is related to our work is physics-informed DeepONet \cite{RN2054} which aim to reduce the reliance of the network on large training datasets by regularizing the training process based on the known PDE system. 

Integral neural operators, first introduced in \cite{RN1508,RN1509}, express the solution operator as an integral operator of Green’s function. Their network primarily consist of lifting, iterative kernel integration, and projection layers. To accelerate the kernel evaluations and increase efficiency, FNO \cite{RN1257} was developed where the integral kernel is parameterized in the Fourier space where the fast Fourier transform (FFT) and its inverse are used to accelerate the computations. Similar to DeepONets, FNO has been extended in multiple directions to better capture spatial signal behaviors \cite{RN2059} or to build a physics-informed neural operator (PINO) that leverages the physics \cite{RN2007} to reduce the reliance on training data.

To build more flexible neural operators that accommodate irregular geometries and sampling grids, \cite{RN2010} introduces a coordinate-based model that leverages implicit neural representations to encode functions into low-dimensional latent spaces and, in turn, infer mappings between function representations in the latent space. Another example is GNOT \cite{RN2011} which is a transformer-based model that leverages attention and geometric gating mechanisms to handle irregular data grids and multiple input functions. Multi-grid techniques which are ubiquitously used in numerical solvers have also been combined with deep NNs to increase the accuracy and flexibility of neural operators \cite{RN2060,RN2061,RN2053}. 

\noindent \textbf{Operator-valued Kernels:} 
Functional data analysis (FDA) \cite{ramsay1991some} also deals with learning mappings between function spaces and has been extensively studied in the statistics literature as a natural extension of multi-response (aka multi-task or multi-output function) learning problems \cite{micchelli2005learning,breiman1997predicting,evgeniou2004regularized,argyriou2008convex,alvarez2012kernels}. In this context, the focus has primary been on the use of RKHS theory which naturally enables the construction of optimal recovery maps between function spaces via the representer formula \cite{owhadi2023ideas,chen2021solving,owhadi2022computational}. 
In the context of learning PDE operators, recently, \cite{batlle2024kernel} introduced a framework based on the theory of operator-valued RKHS and GPs where the reconstruction maps (corresponding to the observation operators) as well as the finite-dimensional map connecting $u^p$ and $v^q$ are all formulated as optimal recovery problems whose solution can be identified by the representer formula. Motivated by this work and FNO, kernel neural operators \cite{lowery2024kernel} are recently developed that use parameterized, closed-form, finitely-smooth, and compactly-supported kernels with trainable sparsity parameters in integral operators. A more technical review of \cite{baydin2018automatic} and its relation to this article is provided in \Cref{subsec comparison}.

\noindent \textbf{GP Regression:} 
The mapping in \Cref{eq function learning problem} is between finite-dimensional Euclidean spaces and hence can be numerically approximated via GPs which have long been used for probabilistic regression \cite{RN1275, RN1957, RN1479,RN1838,RN1845,RN1935}. 
Over the past few years, GPs have attracted some attention for solving nonlinear PDE systems. 
The majority of existing works \cite{RN1873, RN1919, RN1890} employ zero-mean GPs and with this choice, solving a PDE system amounts to designing the GP's kernel whose parameters are obtained via either MLE or a regularized MLE where the penalty term quantifies the GP's error in satisfying the PDE system.
Using zero-mean GPs, \cite{RN1886} casts solving nonlinear PDEs as an optimal recovery problem whose loss function is derived based on the PDE system and aims to estimate the solution at a finite number of interior nodes in the domain. Once these values are estimated, the PDE solution is approximated anywhere in the domain via kernel regression. The scalability of this approach can be increased using sparse GPs \cite{RN1878,RN1883,RN1890}. 
GPs with non-zero means have been employed for solving PDEs in a data-driven manner \cite{wang2023discovery} but \cite{mora2024neural} introduced the first robust mechanism for solving nonlinear PDE systems via NN-mean GPs for both forward and inverse problems. This latter work serves as a foundation for the current article. 

\subsection{Contributions and Article Outline}
We summarize our main contributions as follows:
\begin{itemize}
    \item We introduce a hybrid GP/NN-based operator learning framework. Our approach incorporates both zero-mean and NN-mean GPs both of which provide very competitive performance. The latter option, in particular, leverages the strengths of both kernel methods and neural operators such as DeepONet and FNO. 
    \item We consider both data-driven and physics-informed operator learning scenarios. In the former case, we use MLE for parameter optimization and set the mean function to be either zero or a neural operator. In the latter case, we only use NN-mean GPs whose parameters are estimated via a weighted combination of MLE and PDE residuals. 
    \item We address the computational costs and ill-conditioning issues associated with the covariance matrix of GPs by $(1)$ leveraging specific kernel structures and $(2)$ proper initialization of the kernel parameters that are either trained (for zero-mean GPs) or frozen (for NN-mean GPs).
    \item We introduce the first \textit{zero-shot} mechanism for operator learning based on zero-mean GPs whose kernel parameters are properly initialized. 
    \item We test our approach on single- and multi-output operator learning problems and demonstrate how the prediction accuracy is affected by $(1)$ replacing MSE via MLE in data-driven operator learning, $(2)$ augmenting MLE with physics in the case of NN-mean GPs, and $(3)$ using the trained models for extrapolation. 
\end{itemize}

The rest of the paper is organized as follows. We introduce our framework in \Cref{sec method} where we consider both data-driven and physics-informed scenarios in \Cref{subsec data operator learning,subsec physics operator learning}, respectively. Details on the initialization of the kernel parameters, stability of the covariance matrix, and inference costs are also included in \Cref{sec method}. We compare our approach against competing methods in \Cref{sec results} using single- and multi-output operators with and without incorporating physics into the operator learning problem. Concluding remarks and future research directions are provided in \Cref{sec conclusions}.
    \section{Proposed Framework for Operator Learning} \label{sec method}
We first introduce our approach in the context of data-driven operator learning in \Cref{subsec data operator learning} and then consider the physics-informed version in \Cref{subsec physics operator learning} to enable augmenting the training data with physical laws. The data-driven scenario in \Cref{subsec data operator learning} can leverage either zero-mean or non-zero mean GPs while the physics-informed one in \Cref{subsec physics operator learning} relies on non-zero mean GPs. 
To ensure numerical stability and computational efficiency, in \Cref{subsec stability and initialization} we study the kernel parameters and provide general guidance for properly initializing them and potentially updating them in zero-mean and NN-mean GPs.
We conclude this section by comparing our approach to the optimal recovery-based approach of \cite{batlle2024kernel} and then commenting on its inference complexity in \Cref{subsec comparison,subsec complexity}, respectively.

\subsection{Data-driven Operator Learning} \label{subsec data operator learning}
\subsubsection{Single-output Operators} \label{subsubsec scalar valued operators}
In \Cref{subsec method summary} we transformed the operator learning problem into an equivalent regression one with the goal of learning $\mathGtildedaggerp: \real^n \times \Omega_v \rightarrow \real$. To approximate $\mathGtildedaggerp$ we start by placing a GP prior on it:
\begin{equation}
    \mathGtildedaggerp \sim GP\parens{m(\phi(u), y; \theta), c\parens{\brackets{\phi(u), y}, \brackets{\phi(u'), y'}; \beta, \sigma^2}}
    \label{eq gp prior}
\end{equation}
where $\phi(u)$ is the linear observation operator defined in \Cref{eq observation operators}, $m(\phi(u), y; \theta)$ is the prior mean function parameterized by $\theta$, and $c\parens{\brackets{\phi(u), y}, \brackets{\phi(u'), y'}; \beta, \sigma^2}$ is a kernel with hyperparameters $\beta$ and $\sigma^2$. We elaborate on the choice of the mean function later in this section. As for the kernel, there are many choices available but in this paper we employ the Gaussian or Mat\'ern covariance functions defined as:
\begin{subequations} 
    \begin{equation} 
        c(x, x^{\prime}; \beta, \sigma^2)=
        \sigma^2 \exp \braces{-\parens{x-x'}^T \text{diag}\parens{\beta} \parens{x-x'}} =
        \sigma^2 \exp \braces{-\sum_{i=1}^{d_x} \beta_i(x_i - x'_i)^2}
        \label{eq gaussian kernel}
    \end{equation}
    \begin{equation} 
        c(x, x^{\prime}; \beta, \sigma^2) = \sigma^2\frac{2^{1-\nu}}{\Gamma(\nu)} 
        \parens{\sqrt{2\nu} \parens{x-x'}^T \text{diag}\parens{\beta} \parens{x-x'}}^{\nu}
        K_{\nu} \parens{\sqrt{2\nu} \parens{x-x'}^T \text{diag}\parens{\beta} \parens{x-x'}}
        \label{eq matern kernel}
    \end{equation}
    \label{eq all kernels}
\end{subequations} 
where $x$ denotes the feature vector with length $d_x$, $\beta$ are the length-scale parameters, $\sigma^2$ is the process variance, $\nu \in \{\frac{1}{2},\frac{3}{2}, \frac{5}{2} \}$, $K_{\nu}$ is the modified Bessel function of the second kind, and $\Gamma$ is the gamma function. The inductive bias that the kernels in \Cref{eq all kernels} encode into the learning process is that close-by inputs $x$ and $x'$ have similar (i.e., correlated) output values. The degree of this correlation also depends on $\beta$ whose magnitude quantifies the pace at which the correlations die out as the distance between $x$ and $x'$ increases.

Predicting $\mathGdagger(u^*)(y^*)$ for an unseen input function $u^*$ at unseen output location $y^*$ requires the estimation of $\theta$, $\sigma^2$, and $\beta$. To this end, we combine the training data $\mathD :=\{\phi(u_i), \psi(v_i)\}_{i=1}^N$ with maximum likelihood estimation. To accommodate our regression problem, we reorganize $\mathD$ as $\{(\phi(u_i), y_j), [\psi(v_i)]_j) \}_{i,j=1}^{N,q}$ where $\phi(u_i) \in \real^p$ and $y_j \in \Omega_v$ denote an input pair whose corresponding scalar output is $[\psi(v_i)]_j \in \real$.
For notational convenience, we denote the inputs and outputs of this dataset by $\Xb=\braces{\phi(u_i),y_j}_{i,j=1}^{N,p}$ and $\vb = \braces{[\psi(v_i)]_j}_{i,j=1}^{N,p}$, respectively. 
The MLE process involves the following maximization problem:
\begin{equation} 
    \begin{split}
        \brackets{\widehat \theta, \widehat \beta, \widehat \sigma^2} = 
        \underset{\theta \in \Theta, \beta \in \mathcal{B}, \sigma^2 \in \real^+}{\operatorname{argmax}} \hspace{2mm} 
        (2 \pi)^{-Nq/2} \left|\Cb \right|^{-\frac{1}{2}} 
        \exp \left\{\frac{-1}{2}\parens{\vb - \mb}^T \Cb^{-1}\parens{\vb - \mb }\right\},
    \end{split}
    \label{eq mle}
\end{equation}
or equivalently:
\begin{equation} 
    \begin{split}
        \brackets{\widehat \theta, \widehat \beta, \widehat \sigma^2} = 
        \underset{\theta \in \Theta, \beta \in \mathcal{B}, \sigma^2 \in \real^+}{\operatorname{argmin}} \hspace{2mm}
        \frac{1}{2} \log{\bars{\Cb}} + \frac{1}{2}\parens{\vb - \mb}^T \Cb^{-1}\parens{\vb - \mb },
    \end{split}
    \label{eq nll}
\end{equation}
where $\Cb=c\parens{\Xb, \Xb; \beta, \sigma^2}$ is the $Nq \times Nq$ symmetric positive-definite covariance matrix, $\mb = m\parens{\Xb; \theta}$ is the $Nq \times 1$ vector of mean function evaluations at $\Xb$, $\Theta$ and $\mathcal{B}$ denote the search spaces for $\theta$ and $\beta$, respectively, and $\log\bars{\Cb}$ denotes the log-determinant of $\Cb$. The $\vb - \mb$ vector quantifies the residuals of the mean function in reproducing the training outputs and is discussed further in \Cref{subsec inference}.

The minimization problem in \Cref{eq nll} is almost always solved with a gradient-based optimization technique where the parameters are first initialized and then iteratively updated until a convergence metric (e.g., maximum number of iterations or function evaluations) is met. The overall computational cost of this iterative process is dominated by the repeated inversion of the covariance matrix $\Cb$. Specifically, the computational complexity associated with $\Cb^{-1}$ is $\mathO\parens{\parens{Nq}^3}$ which poses a major challenge in operator learning problems where $N$ is typically in the order of thousands. Even with relatively small $N$, inverting $\Cb$ can still be very costly for high-resolution data where $q$ is large. 

To reduce the computational costs of inverting $\Cb$, we exploit the structure of the data where we assume that the discretization of $u$ and the locations where $v$ is observed are fixed across the $N$ training samples. This assumption is quite reasonable in many operator learning problems and allows us to leverage the Kronecker product properties to bypass the construction and inversion of $\Cb$. To this end, we adopt a separable kernel of the form:
\begin{equation}
    c([\phi(u), y], [\phi(u'), y']; \beta, \sigma^2) = 
    c_{\phi}(\phi(u),\phi(u^{\prime}); \beta_{\phi}, \sigma^2_\phi)
    c_y(y,y^{\prime}; \beta_y, \sigma^2_y)
    \label{eq kernel separate}
\end{equation}
where we use \Cref{eq all kernels} for both kernels on the right-hand side. Without loss of generality, we assume $\sigma^2_y = 1$ hereafter\footnote{Alternatively, we could have also assumed $\sigma^2_\phi = 1$ and estimate $\sigma^2_y$ instead, leading to the same result.}. Using this formulation and the assumption on the data structure we rewrite the covariance matrix in \Cref{eq nll} as:
\begin{equation}
    \Cb= \Cphib \otimes \CYb,
    \label{eq kronecker}
\end{equation}
where $\Cphib = c_{\phi}(\Ub, \Ub; \beta_{\phi}, \sigma^2_\phi)$ and $\CYb = c_y(\Yb, \Yb; \beta_y)$ with $\Ub = \braces{\phi(u_1),\dots,\phi(u_N)}$ and $\Yb = \braces{y_1,\dots,y_q}$, $\beta = \brackets{\beta_\phi, \beta_y}$. In this equation, $\Cphib$ and $\CYb$ model the correlations among the discretized input functions and the locations where the output function is observed, respectively.  
Using the Kronecker product in \Cref{eq kronecker} we can simplify three of the computationally expensive operations in \Cref{eq nll}:
\begin{subequations}
    \begin{align}
        & \Cb^{-1} = \Cphib^{-1} \otimes \CYb^{-1}, 
        \label{eq kronecker property 1} \\
        & \Cb^{-1} \parens{\vb - \mb} = \vecOp{\parens{\CYb^{-1}\parens{\Vb - \Mb} \Cphib^{-1}}}
        \label{eq kronecker property 2} \\
        &\bars{\Cb} = \bars{\Cphib}^q \bars{\CYb}^N, 
        \label{eq kronecker property 3}
    \end{align}
    \label{eq kronecker properties}        
\end{subequations}
where $\Vb$ and $\Mb$ are formed by reshaping $\vb$ and $\mb$ to $q \times N$ matrices, respectively, and $\vecOp{\parens{\cdot}}$ denotes the vectorization operation. Using \Cref{eq kronecker properties} we now rewrite \Cref{eq nll} as:
\begin{equation} 
    \begin{aligned}
        \brackets{\widehat \theta, \widehat \beta, \widehat \sigma^2_\phi} &= 
        \underset{\theta \in \Theta, \beta \in \mathcal{B}, \sigma^2_\phi \in \real^+}{\operatorname{argmin}} \hspace{2mm} \mathL_{\text{MLE}} (\theta, \beta, \sigma^2_\phi) \\
        &= \underset{\theta \in \Theta, \beta \in \mathcal{B}, \sigma^2_\phi \in \real^+}{\operatorname{argmin}} \hspace{2mm} 
        \log{\parens{\bars{\Cphib}^q \bars{\CYb}^N}} + \parens{\vb - \mb}^T \vecOp{\parens{\CYb^{-1}\parens{\Vb - \Mb} \Cphib^{-1}}},
    \end{aligned}
    \label{eq nll kronecker}
\end{equation}
which can be used to simultaneously estimate the parameters of the mean and covariance functions of the GP. 

We highlight that the optimization problems in \Cref{eq nll,eq nll kronecker} both correspond to our generic GP-based operator learning framework except that the latter one exploits the data structure to decrease the computational costs and memory demands. If the locations where the output function is observed or the discretization of the input function vary across the samples, \Cref{eq nll kronecker} can still be used by either pre-processing the data (which may introduce some additional errors) or sparse GPs. 

For a zero-mean GP, \Cref{eq nll kronecker} simplifies to:
\begin{equation} 
    \begin{split}
        \brackets{\widehat \beta, \widehat \sigma^2_\phi} &= 
        \underset{\beta \in \mathcal{B}, \sigma^2_\phi \in \real^+}{\operatorname{argmin}} \hspace{2mm} \mathL_{\text{0}} (\beta, \sigma^2_\phi)\\
        &= \underset{\beta \in \mathcal{B}, \sigma^2_\phi \in \real^+}{\operatorname{argmin}} \hspace{2mm} 
        q\log{\parens{\bars{\Cphib}}}+N\log{\parens{\bars{\CYb}}}
        + \vb^T \vecOp{\parens{\CYb^{-1} \Vb \Cphib^{-1}}},
    \end{split}
    \label{eq nll kronecker zero mean}
\end{equation}
which can be solved via a first-order gradient-based optimizer (e.g., Adam) that leverages automatic differentiation to obtain the gradients of $\mathL_{\text{0}} (\beta, \sigma^2_\phi)$ with respect to $\beta$ and $\sigma^2_\phi$. 
In typical GP regression problems, it is common practice to initialize gradient-based optimizers multiple times to avoid local optimality. This process substantially increases the cost of minimizing $\mathL_{\text{0}} (\beta, \sigma^2_\phi)$ due to the high costs of repeatedly constructing and inverting $\Cphib$ and $\CYb$ to calculate $\mathL_{\text{0}} (\beta, \sigma^2_\phi)$ and its gradients. We address this computational issue in \Cref{subsec stability and initialization} where we show that $\beta_y$ and $\beta_\phi$ can be, respectively, fixed and initialized to some judiciously chosen values to, in turn, optimize $\mathL_{\text{0}} (\beta_\phi, \sigma^2_\phi)$ only once.

For a GP whose mean function is parameterized with a deep NN (e.g., FNO or DeepONet), minimizing $\mathL (\theta, \beta, \sigma^2_\phi)$ can be prohibitively costly and memory intensive (even if $\beta$ are initialized well and the optimization problem in \Cref{eq nll kronecker} is solved only once) since differentiating $\mathL (\theta, \beta, \sigma^2_\phi)$ with respect to $\theta$ involves large matrices. Moreover, we show in \Cref{subsec stability and initialization} that the simultaneous estimation of $\theta$ and $\beta$ renders the covariance matrices ill-conditioned. As detailed in \Cref{subsec stability and initialization}, we address these computational issues by fixing $\beta$ and $\sigma^2_\phi$ to some values that ensure $(1)$ the covariance matrices are numerically stable, and $(2)$ the posterior distributions in \Cref{subsec inference} regress the training data. 
Fixing $\beta$ and $\sigma^2_\phi$ makes the first term in \Cref{eq nll kronecker} a constant and hence we can estimate $\theta$ via:
\begin{equation} 
    \begin{split}
        \widehat \theta = \underset{\theta \in \Theta}{\operatorname{argmin}} \hspace{2mm} \mathL_{\text{nn}} (\theta)
        = \underset{\theta \in \Theta}{\operatorname{argmin}} \hspace{2mm} 
        \parens{\vb - \mb}^T \vecOp{\parens{\CYb^{-1}\parens{\Vb - \Mb} \Cphib^{-1}}}.
    \end{split}
    \label{eq nll kronecker nn mean}
\end{equation}
where $\mb$ and $\Mb$ are the only terms that depend on $\theta$, i.e., we can construct $\CYb^{-1}$ and $\Cphib^{-1}$ once and store them so that evaluating $\mathL_{\text{nn}} (\theta)$ and its gradients only relies on matrix multiplications. 

We notice in \Cref{eq nll kronecker nn mean} that if $\CYb$ and $\Cphib$ are identity matrices (i.e., if the correlations among different $u^p$ and different query points $y$ are eliminated), $\mathL_{\text{nn}} (\theta)$ becomes the mean squared error (MSE) loss function that is typically used in operator learning. We demonstrate the benefits of using $\mathL_{\text{nn}} (\theta)$ instead of MSE in \Cref{sec results}.

\subsubsection{Multi-output Operators} \label{subsubsec vectorvalued operators}
Consider now the following multi-output operator 
\begin{equation}
    \mathGdagger: \mathU \rightarrow \prod_{s=1}^{S} \mathV^s
    \label{eq vector valued operator learning problem}
\end{equation}
where $S$ is the total number of (possibly related) output functions. In this case, our goal is to approximate $\mathGdagger$ given the training dataset $\mathD^S := \{s, \phi(u_i), \psi(v^s_i)\}_{i=1}^{N}$ where $\{u_i,v^s_i\}_{i=1}^{N}$ are elements from $\mathU \times \mathV^s$ with $s=\braces{1,...,S}$.
Analogously to \Cref{subsec method summary}, we cast the operator learning problem in \Cref{eq vector valued operator learning problem} as the following regression problem:
\begin{equation}
    \mathGtildedaggerp: \real^p \times \mathVstar \rightarrow \real^S.
    \label{eq vector valued function learning problem}
\end{equation}
where now $\mathGtildedaggerp$ is a \textit{multi-output} function and $\mathVstar$ denotes the dual of the product space $\prod_{s=1}^{S} \mathV^s$. To approximate $\mathGtildedaggerp$, we start by placing a multi-task or $S-$dimensional GP prior \cite{bonilla2007multi,RN287} on it:
\begin{equation}
    \mathGtildedaggerp \sim GP\parens{m(\phi(u), y; \theta)(s), c\parens{\brackets{s, \phi(u), y}, \brackets{s, \phi(u), y}^\prime; \beta, \sigma^2}}
    \label{eq multi task gp prior}
\end{equation}
where $m(\phi(u), y; \theta)$ is the $S-$dimensional mean function of the GP. 

The covariance matrix of the training data cannot be used in training unless $c\parens{\brackets{s, \phi(u), y}, \brackets{s, \phi(u), y}^\prime; \beta, \sigma^2}$ possesses certain features that eliminate the need to build and invert the $NSq \times NSq$ covariance matrix. 
Hence, we invoke the assumptions made on the data structure in \Cref{subsubsec scalar valued operators} and also presume that all the outputs are observed at the same collocation points (CPs). 
Following these assumptions, we formulate the kernel in \Cref{eq multi task gp prior} as:
\begin{equation} 
    \begin{split}
        c\parens{\brackets{s, \phi(u), y}, \brackets{s, \phi(u), y}^\prime; \beta, \sigma^2} = 
        c_s(s,s^{\prime}; \beta_s, \sigma^2_s) 
        c_y(y,y^{\prime}; \beta_y, \sigma^2_y)
        c_{\phi}(\phi(u),\phi(u^{\prime}); \beta_{\phi}, \sigma^2_\phi) ,
    \end{split}
    \label{eq multi task kernel}
\end{equation}
where $c_y(y,y^{\prime}; \beta_y, \sigma^2_y)$ and $c_{\phi}(\phi(u),\phi(u^{\prime}); \beta_{\phi}, \sigma^2_\phi)$ are standard kernels such as the Gaussian or Matérn defined in \Cref{eq all kernels}, $\beta = [\beta_s, \beta_y, \beta_{\phi}]$, and $c_s(s,s^{\prime}; \beta_s, \sigma^2_s)$ captures the average correlations among the responses. Once again, and without loss of generality, we assume $\sigma^2_y = \sigma^2_s = 1$ and hence only consider $\sigma^2_\phi$. The kernel in \Cref{eq multi task kernel} has a separable structure and is stationary since all the kernels on the right-hand side are stationary. 

To further simplify the computations we set $c_s(s,s^{\prime}; \beta_s)$ to the identity matrix, that is:
\begin{equation} 
    \begin{split}
    c_s(s,s^{\prime}) = \mathbbm{1}\braces{s==s^\prime},
    \end{split}
    \label{eq task kernel}
\end{equation}
where $\mathbbm{1}\{\cdot\}$ returns $1/0$ if the enclosed statement is true/false. With this choice, the correlations between the different responses are either neglected or captured by the mean function in zero-mean and NN-mean GPs, respectively. 

To estimate $\theta, \beta$, and $\sigma_\phi^2$ we use the training dataset $\mathD^S$ to maximize the likelihood function. We rearrange $\mathD^S$ as $\{(s,\phi(u_i), y_j), [\psi(v^s_i)]_j \}_{i,j,s=1}^{N,q,S}$ and denote its inputs and outputs via $\Xb=\braces{\phi(u_i),y_j}_{i,j=1}^{N,q}$ and $\vb = \braces{[\psi(v^s_i)]_j}_{i,j,s=1}^{N,q,S}$, respectively, where we have used the assumptions that $\Xb$ do not depend on $s$. The MLE process for our multi-output regression task involves the following minimization problem:
\begin{equation}
    \begin{aligned}
        \brackets{\widehat \theta, \widehat \beta, \widehat \sigma^2_\phi} &= 
        \underset{\theta \in \Theta, \beta \in \mathcal{B},\sigma^2_\phi \in \real^+}{\operatorname{argmin}} \hspace{2mm} \frac{1}{2} 
        \log{\bars{I \otimes \Cb}} + \frac{1}{2}\parens{\vb - \mb}^T (I \otimes \Cb)^{-1}\parens{\vb - \mb } \\
        &= \underset{\theta \in \Theta, \beta \in \mathcal{B}, \sigma^2_\phi \in \real^+}{\operatorname{argmin}} \hspace{2mm} \frac{1}{2} 
        \log{\bars{\Cb}^S} + \frac{1}{2}\parens{\vb - \mb}^T \parens{I \otimes \Cb^{-1}}\parens{\vb - \mb },
    \end{aligned}
    \label{eq multitask mle}
\end{equation}
where $I$ is the $S\times S$ identity matrix, $\Cb=\Cphib \otimes \CYb$ is the $Nq \times Nq$ symmetric positive-definite covariance matrix, and $\mb$ is a column vector of length $NqS$ that corresponds to the evaluations of the mean function for all the outputs, samples, and query points.

Similar to the previous section, we leverage the properties of the Kronecker product in \Cref{eq kronecker properties} to simplify \Cref{eq multitask mle} to:
\begin{equation} 
    \begin{split}
        \brackets{\widehat \theta, \widehat \beta, \widehat \sigma^2_\phi} = &
        \underset{\theta \in \Theta, \beta \in \mathcal{B},\sigma^2_\phi \in \real^+}{\operatorname{argmin}} \hspace{2mm} \frac{1}{2} \log{\parens{\bars{\Cphib}^{qS} \bars{\CYb}^{NS}}} \\
        &+ \frac{1}{2} \sum_{s=1}^S
        \parens{\vb^s - \mb^s}^T \vecOp{\parens{\CYb^{-1}\parens{\Vb^s - \Mb^s} \Cphib^{-1}}},
    \end{split}
    \label{eq multi task nll kronecker}
\end{equation}
where $\Vb^s$ and $\Mb^s$ are formed by reshaping $\vb^s$ and $\mb^s$ to $q \times N$ matrices, respectively, $\vb^s = \braces{[\psi(v^s_i)]_j}_{i,j}^{N,p}$, and $\mb^s = m\parens{s,\Xb; \theta}$ denotes the evaluations of the mean function for response $s$.

Parameter estimation now follows as in \Cref{subsubsec scalar valued operators}. Specifically, for a zero-mean GP, \Cref{eq multi task nll kronecker} simplifies to:
\begin{equation} 
    \begin{split}
        \brackets{\widehat \beta, \widehat \sigma^2_\phi} &= 
        \underset{\beta \in \mathcal{B},\sigma^2_\phi \in \real^+}{\operatorname{argmin}} \hspace{2mm} \mathL_{\text{0v}} (\beta, \sigma^2_\phi) \\
         & = \underset{\beta \in \mathcal{B}, \sigma^2_\phi \in \real^+}{\operatorname{argmin}} \hspace{2mm} \frac{1}{2} \log{\parens{\bars{\Cphib}^{qS} \bars{\CYb}^{NS}}} + \frac{1}{2} \sum_{s=1}^S
        \parens{\vb^s}^T \vecOp{\parens{\CYb^{-1}\Vb^s \Cphib^{-1}}}.
    \end{split}
    \label{eq multi task nll kronecker zero-mean}
\end{equation}
where we can fix $\beta_\phi$ similar to the scalar case in \Cref{subsubsec scalar valued operators}. For an NN-mean GP, we fix the kernel hyperparameters which makes the first term on the right-hand side of \Cref{eq multi task nll kronecker} to become a constant. Hence, we can estimate $\theta$ via:
\begin{equation} 
    \widehat \theta = \underset{\theta \in \Theta}{\operatorname{argmin}} \hspace{2mm} \mathL_{\text{nnv}} (\theta)
    = \underset{\theta \in \Theta}{\operatorname{argmin}} \hspace{2mm} \frac{1}{2} \sum_{s=1}^S
    \parens{\vb^s - \mb^s}^T \vecOp{\parens{\CYb^{-1}\parens{\Vb^s - \Mb^s} \Cphib^{-1}}}.
    \label{eq multitask nll kronecker non-zero-mean}
\end{equation}

\subsubsection{Inference} \label{subsec inference}
Once the parameters of the mean and covariance functions are estimated, we obtain the predictive response for a test input function $u^*$ at the query point $y^*$ by computing the expected value of the posterior distribution of the GP conditioned on the training data. That is:
\begin{equation} 
    \begin{split}
        \eta \parens{\phi(u^*),y^*; \widehat\theta, \widehat\beta, \widehat\sigma_\phi^2} &:= \E[\mathGtildedaggerp(\phi(u^*),\delta_{y^*}) | \mathD] \\
        &= m(\phi(u^*), y^* ; \widehat{\theta}) + c\parens{\brackets{\phi(u^*), y^*}, \Xb; \widehat{\beta}, \widehat \sigma_\phi^2} \boldsymbol{C}^{-1}(\vb-\mb)\\
        &= m(\phi(u^*), y^* ; \widehat{\theta}) + \\
        & \quad \brackets{c_{\phi}\parens{\phi(u^*), \Ub; \widehat{\beta}_\phi, \widehat \sigma^2_\phi} \otimes c_y\parens{y^*, \Yb; \widehat{\beta}_y}} \vecOp{\parens{\CYb^{-1}\parens{\Vb - \Mb} \Cphib^{-1}}} \\
        &=  m(\phi(u^*), y^* ; \widehat{\theta}) + \\
        & \quad \vecOp{\parens{c_y\parens{y^*, \Yb; \widehat{\beta}_y} \CYb^{-1}\parens{\Vb - \Mb} \Cphib^{-1} c_{\phi}\parens{\Ub, \phi(u^*); \widehat{\beta}_\phi, \widehat \sigma^2_\phi} }},
    \end{split}
    \label{eq expected value posterior one sample}
\end{equation}
where we have leveraged the Kronecker product properties in \Cref{eq kronecker properties} to simplify the matrix multiplications. \Cref{eq expected value posterior one sample} accommodates batch computations for accelerated prediction of the output field at the $m$ points $\Ystarb = \braces{y_1^*, \dots, y_m^*}$:
\begin{equation}
    \begin{split}
        \eta \parens{\phi(u^*),\Ystarb; \widehat\theta, \widehat\beta, \widehat \sigma^2_\phi} &= 
 m(\phi(u^*), \Ystarb ; \widehat{\theta}) + \\
        &\vecOp{\parens{c_y\parens{\Ystarb, \Yb; \widehat{\beta}_y} \CYb^{-1}\parens{\Vb - \Mb} \Cphib^{-1} c_{\phi}\parens{\Ub, \phi(u^*); \widehat{\beta}_\phi, \widehat \sigma^2_\phi} }}    
    \end{split}
    \label{eq expected value posterior}
\end{equation}
where $\Ub = \braces{\phi(u_1),\dots,\phi(u_N)}$ and $\Yb = \braces{y_1,\dots,y_q}$ as defined previously. In the case of multi-output operators, our predictor for output $s$ is:
\begin{equation}
    \begin{aligned}
        \eta \parens{s, \phi(u^*),\Ystarb; \widehat\theta, \widehat\beta, \widehat\sigma^2_\phi} &= m(\phi(u^*), \Ystarb ; \widehat{\theta})(s) \\
        &+ \vecOp{\parens{c_y(\Ystarb, \Yb; \widehat{\beta}_y) \CYb^{-1}\parens{\Vb^s - \Mb^s} \Cphib^{-1} c_{\phi}(\Ub,\phi(u^*); \widehat{\beta}_\phi, \widehat \sigma^2_\phi) }}.    
    \end{aligned}
    \label{eq expected value posterior vector valued}
\end{equation}
For a zero-mean GP, \Cref{eq expected value posterior one sample,eq expected value posterior,eq expected value posterior vector valued} can be simplified by removing the mean function and the associated matrices $\Mb$ and $\Mb^s$.

The predictor in \Cref{eq expected value posterior one sample} is a differentiable function with respect to $y^*$. We use this property in \Cref{subsec physics operator learning} for physics-informed operator learning where the training process leverages the known differential operators that govern the problem. This predictor also interpolates the training data which is a feature that most neural operators do not provide. To see this, we use \Cref{eq expected value posterior} to predict the output function at $\Yb$ for the $N$ input functions $\Ub$ that are observed during training:
\begin{equation}
    \begin{aligned}
        \eta\parens{\Ub,\Yb; \widehat\theta, \widehat\beta, \widehat\sigma^2_\phi} &= m(\Ub, \Yb ; \widehat{\theta}) + 
        \vecOp{\parens{c_y(\Yb, \Yb; \widehat{\beta}_y) \CYb^{-1}\parens{\Vb - \Mb} \Cphib^{-1} c_{\phi}(\Ub,  \Ub; \widehat{\beta}_\phi, \widehat\sigma^2_\phi) }} \\
        & = \mb + \vecOp{\parens{I_y \parens{\Vb - \Mb} I_{\phi}}} 
        = \mb + \vb - \mb = \vb.
    \end{aligned}
    \label{eq expected value posterior on train samples}
\end{equation}
Lastly, we highlight that both terms on the right-hand side of \Cref{eq expected value posterior one sample} capture the correlations not only in $\mathU$, but also in the support of $\mathV$, \ie $\Omega_v$. That is, our predictor leverages the strengths of both neural operators and kernel methods for operator learning.

\subsection{Physics-informed Operator Learning} \label{subsec physics operator learning}
Consider the general differential operator $\mathT$ acting upon $u \in \mathU$ and $v \in \mathV$ such that $\mathT: \mathU \times \mathV \rightarrow \mathO$, where $\mathO$ is the null space, $u$ is the input function, and $v$ denotes the unknown solution. Then, a general family of (possibly nonlinear) time-dependent PDEs can be expressed as:
\begin{subequations}
    \begin{align}
        &\mathT(u,v) = 0 \quad \text{in } \Omega_v \times (0, \infty) ,
        \label{eq general PDE} \\
        &v = f \quad \text{on } \partial \Omega_v \times (0, \infty),
        \label{eq general BC} \\
        &v = g \quad \text{in } \bar{\Omega}_v \times \braces{0}.
        \label{eq general IC}
    \end{align}
    \label{eq general PDE system}        
\end{subequations}
Given \Cref{eq general PDE system} and $N$ pairs of $\{\phi(u_i), \psi(v_i)\}$ that satisfy it, our goal is to approximate the operator $\mathGdagger: \mathU \rightarrow \mathV$. 
In this context, $\mathU$ may represent a family of PDE coefficients $a(y,t):\Omega_v \times (0, \infty) \rightarrow \real$ where $t$ denotes time, boundary conditions $v(y,t):\partial \Omega_v \times (0, \infty) \rightarrow \real$, or initial conditions $v(y,t): \bar{\Omega}_v \times \braces{0} \rightarrow \real$. 

To improve the approximation accuracy of the predictor in \Cref{eq expected value posterior one sample}, we regularize the parameter estimation process via \Cref{eq general PDE system}. To this end, we use a batch of $N_{pi}$ samples with inputs $\braces{\phi(u_i)}_{i=1}^{N_{pi}}$ which may or may not be unique from the $N$ training samples used throughout \Cref{subsec data operator learning}. 
For each of the $N_{pi}$ samples, we consider $n_\text{PDE}$, $n_\text{BC}$, and $n_\text{IC}$ collocation points in the domain, on the boundaries, and at $t=0$, respectively, to obtain the following losses:
\begin{subequations}
    \begin{align}
        &\mathL_{\text{PDE}}(\theta) = \frac{1}{N_{pi} n_\text{PDE}} \sum_{j=1}^{N_{pi}} \sum_{i=1}^{n_\text{PDE}}\parens{\mathT(u_j,\eta(\phi(u_j),\delta_{y_i})}^2, 
        \label{eq loss pde} \\
        &\mathL_{\text{BC}}(\theta) = \frac{1}{N_{pi} n_\text{BC}} \sum_{j=1}^{N_{pi}} \sum_{i=1}^{n_\text{BC}}\parens{ \eta(\phi(u_j),\delta_{y_i}) - f(y_i)}^2, 
        \label{eq loss bc pde} \\
        &\mathL_{\text{IC}}(\theta) = \frac{1}{N_{pi} n_\text{IC}} \sum_{j=1}^{N_{pi}} \sum_{i=1}^{n_\text{IC}}\parens{ \eta(\phi(u_j),\delta_{y_i}) - g(y_i) }^2,  
        \label{eq loss ic pde}
    \end{align}
    \label{eq loss terms physics}        
\end{subequations}
which quantify the errors of \Cref{eq expected value posterior one sample} in satisfying \Cref{eq general PDE system}. Note that we calculate $\mathL_{\text{BC}}$ and $\mathL_{\text{IC}}(\theta)$ since the $N_{pi}$ samples may be different from the $N$ labeled training samples. But if these two sets of samples have identical ICs/BCs, $\mathL_{\text{BC}}$ and $\mathL_{\text{IC}}(\theta)$ in \Cref{eq loss terms physics} will be zero due to the reproducing property of the predictor, see \Cref{eq expected value posterior on train samples}.

To estimate $\theta$, we combine the loss terms in \Cref{eq loss terms physics} with \Cref{eq nll kronecker nn mean}:
\begin{equation}
   \mathL(\theta) = \mathL_{\text{PDE}}(\theta) + \alpha_{\text{BC}} \mathL_{\text{BC}}(\theta) + \alpha_{\text{IC}} \mathL_{\text{IC}}(\theta) + \alpha_{\text{MLE}} \mathL_{\text{nn}}(\theta),
    \label{eq loss physics data}
\end{equation}
where $\alpha_{\text{BC}}, \alpha_{\text{IC}}$, and $\alpha_{\text{MLE}}$ are weights that ensure none of the loss terms in \Cref{eq loss physics data} dominates the optimization process. Specifically, we choose $\mathL_{\text{PDE}}$ as a reference and assign the unit weight to it. Then, we determine the weights of the other loss terms via an Adam-inspired approach \cite{pgcan2024, wang2021understanding} such that their gradients with respect to $\theta$ have comparable magnitudes at all optimization iterations.
For example, we obtain $\alpha_{\text{BC}}$ at epoch $k+1$ via:
\begin{equation}
    \alpha_{\text{BC}}^{k+1} = (1 - \lambda) \alpha_{\text{BC}}^{k} + \lambda \frac{\max \left| \nabla_\theta \mathL_{\text{PDE}}(\theta) \right|}{\mathrm{mean} \left| \nabla_\theta \mathL_{\text{BC}}(\theta) \right|},
    \label{eq adam inspired}
\end{equation}
where $\mathL_{\text{PDE}}(\theta)$ is chosen as the reference term, $\nabla_\theta \mathL_{\text{BC}}(\theta)$ denotes the gradient of $\mathL_{\text{BC}}(\theta)$ with respect to $\theta$, and $\lambda$ is typically set to $0.9$. 

We conclude this section by highlighting three points. 
First, if the regularized loss function in \Cref{eq loss physics data} involves nonlinear functions of the predictor (i.e., if the PDE system in \Cref{eq general PDE system} is nonlinear), then the posterior in \Cref{eq expected value posterior one sample} is not a Gaussian and the predictor is merely a maximum a posteriori (MAP) estimator. 
Second, for physics-informed operator learning we only use NN-mean GPs with fixed kernel parameters because $(1)$ optimizing the kernel parameters requires the repeated inversion of the covariance matrices which is computationally expensive and prone to ill-conditioning, and $(2)$ our kernels are quite simple and have insufficient parameters to accommodate residual minimization. 
Third, for multi-output operators we replace $\mathL_{\text{nn}}(\theta)$ by $\mathL_{\text{nnv}}(\theta)$ in \Cref{eq multitask nll kronecker non-zero-mean}.

\subsection{Parameter Initialization and Stability} \label{subsec stability and initialization}
For a zero-mean GP we fix $\beta_y$ to some \textit{large} values and only optimize $\beta_\phi$ and $\sigma^2_\phi$ in \Cref{eq nll kronecker zero mean} or \Cref{eq multi task nll kronecker zero-mean}. We rationalize this decision in \Cref{subsubsec dense observations} and then in \Cref{subsubsec high-dim features} we elaborate on our logic for initializing $\beta_\phi$ to some \textit{small} values before they are optimized. For an NN-mean GP, we optimize $\theta$ while $\beta = \brackets{\beta_y, \beta_\phi}$ and $ \sigma^2_\phi$ are all fixed to their initial values which are the same as those for a zero-mean GP. As detailed in \Cref{subsubsec numerical instability} this approach dramatically accelerates the computations while reducing the numerical instabilities. 



\subsubsection{Effect of Observation Operator} \label{subsubsec dense observations}
In many operator learning problems $\psi(v)$ provides dense observations, i.e., $q$ in \Cref{eq observation operators} is a relatively large number. To demonstrate the effect of dense observations on the optimum kernel parameters of a zero-mean GP, we consider the simple task of interpolating $u(x)=\sin \parens{2\pi x}, x \in \brackets{0,1}$ using a zero-mean GP with the Gaussian kernel $c(x,x';\beta, \sigma^2)$. We set $\sigma^2=1$ without loss of generality and sample $N>>1$ points from $u(x)$. Instead of optimizing $\beta$ via MLE, we vary it in the $\brackets{0, 10^6}$ range and for each value we calculate the error of the corresponding conditional predictor in \Cref{eq expected value posterior on train samples} on a large set of test samples. 

\begin{figure}[!h]
    \centering
    \includegraphics[width=1.0\linewidth]{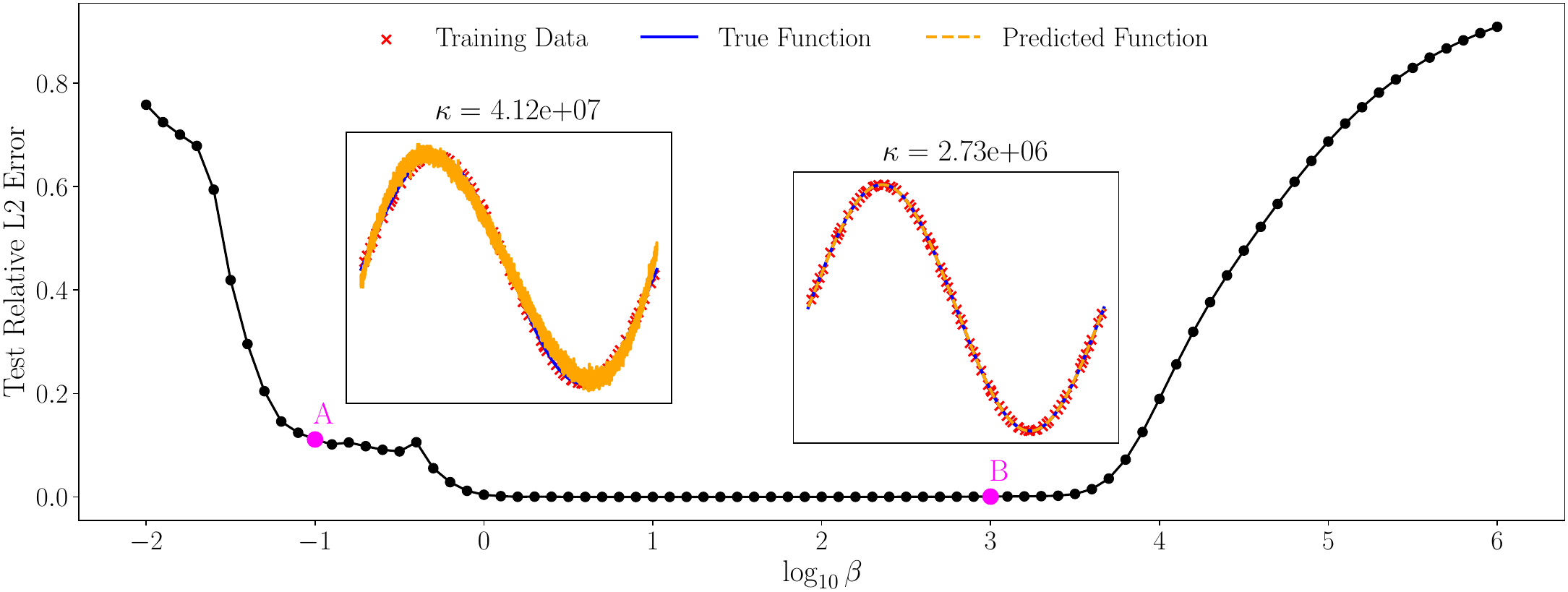}
    \caption{\textbf{Effect of $\beta_y$ on test accuracy with dense observations:} Very small and large $\beta$ values provide poor performance as they cause numerical issues and cannot capture spatial correlations, respectively. However, a relatively large range of values such as $10^3$ provide high accuracy and numerical stability.}
    \label{fig dense observations}
\end{figure}

The results of this study are summarized in \Cref{fig dense observations} where we observe that not all values of $\beta$ result in accurate interpolation of the test samples: $\beta < 1$ renders the covariance matrix ill-conditioned while $\beta>10^4$ prevents the GP from modeling the spatial correlations.
For instance, consider points A and B on the curve which mark $\beta = 10^{-1}$ and $\beta = 10^{3}$, respectively, and the corresponding insets illustrate the predictions of the resulting GPs. We observe in inset A that the prediction curve with $\beta = 10^{-1}$ is quite rough which is due to the ill-conditioning of the GP's covariance matrix. In contrast, for $\beta \in \brackets{10^1, 3.50\times 10^3}$ the resulting GP interpolates unseen data very accurately and provides a smooth prediction curve. Further increasing $\beta$ reduces the prediction accuracy as the posterior mean cannot leverage the spatial correlations at all.

Our studies indicate that the trends in \Cref{fig dense observations} extend to not only higher dimensions, but also the profile of the likelihood function (we further elaborate on the latter point in \Cref{subsubsec high-dim features}). Following these observations, in all of our studies we set $\beta_y$ to large values, \eg $3$, regardless of $d$ (i.e., the dimensionality of the output space $\Omega_v$) and the choice of the GP's mean function (NN or zero). This decision ensures that $C_y$ is numerically stable and the GP can faithfully interpolate the dense observations in $\Omega_v$.



\subsubsection{High-Dimensional Features in GPs} \label{subsubsec high-dim features}
The performance of GPs deteriorates in very high feature spaces \cite{frazier2018tutorial} due to the so-called curse of dimensionality: kernels capture the correlations between the samples based on their distance in the feature space. Most kernels such as those in \Cref{eq all kernels} poorly characterize these correlations in high-dimensional feature spaces as the quantified distances quickly increase and, in turn, render the correlations among samples to become zero. 
To illustrate this, consider the kernel in \Cref{eq gaussian kernel} with $\sigma^2 = 1$. Assuming the same length-scale parameter $\beta$ is used across all $d_x$ dimensions, we can express $c(x, x^\prime; \beta)$ as a function of the distance between two feature vectors $x$ and $x^\prime$:
\begin{equation}
    c(x, x^{\prime}; \beta)=
    \exp \braces{- \beta \bars{\bars{x-x'}}^2},
    \label{eq gaussian distance}
\end{equation}
which can be solved for $\beta$:
\begin{equation}
    \beta = - \frac{\log{c(x, x^{\prime})}}{\bars{\bars{x-x'}}^2}.
    \label{eq beta}
\end{equation}

\begin{figure}[!h]
    \centering
    \includegraphics[width=1.0\linewidth]{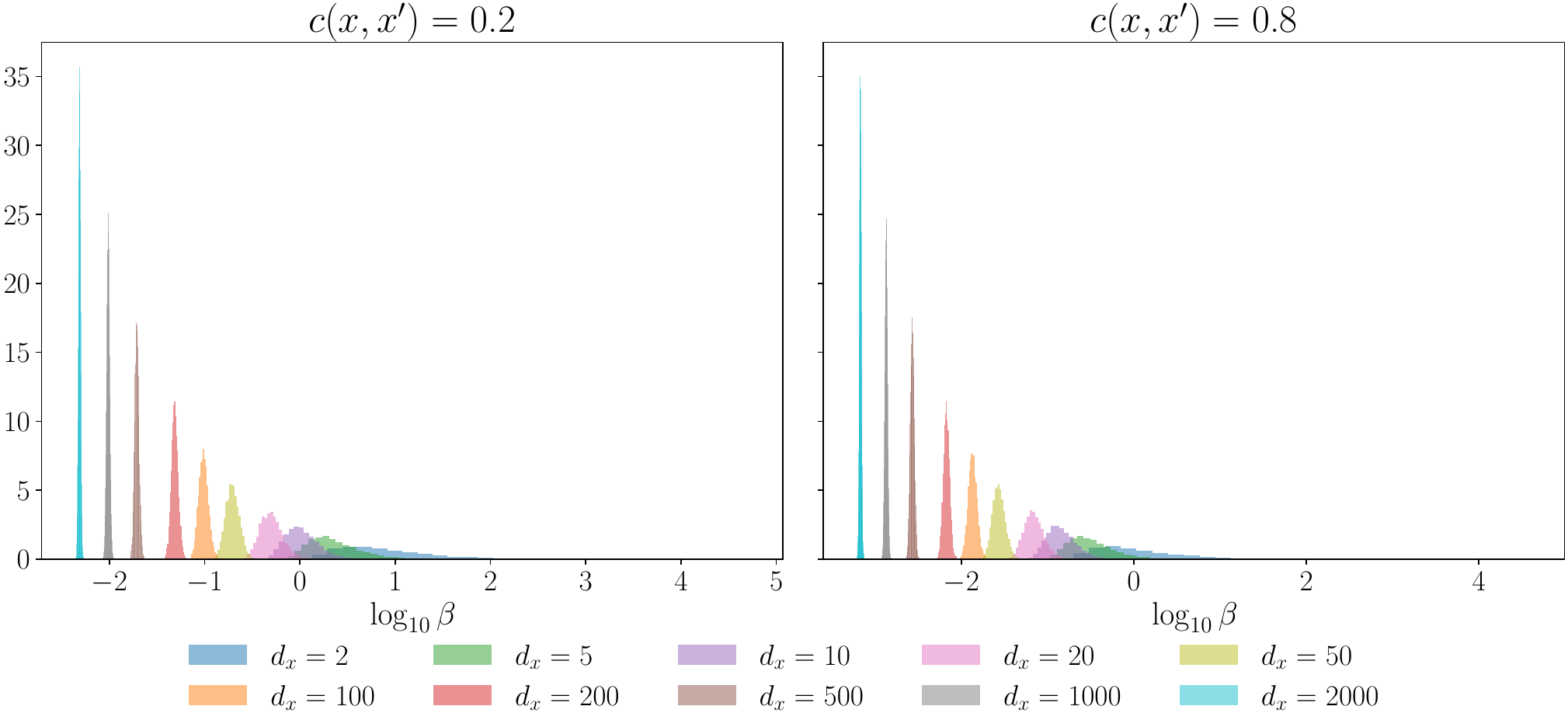}
    \caption{\textbf{Interaction between $\beta$ and feature space dimensionality:} Smaller $\beta$ values are needed to achieve a desired correlation value in high-dimensional feature spaces.}
    \label{fig high-dim histograms}
\end{figure}

To explore the interaction between $d_x$ and $\beta$, we perform the following experiment. Given fixed covariance values $c(x,x^\prime) = 0.2$ and $c(x,x^\prime) = 0.8$, we generate $10,000$ random samples in the $d_x-$dimensional unit hypercube where the norm of each sample is used as the denominator of \Cref{eq beta}. Then, we calculate $\beta$ corresponding to each sample and repeat this process for various values of $d_x$. The corresponding histograms are provided in \Cref{fig high-dim histograms} and demonstrate that as $d_x$ increases, the range of $\beta$ values that achieve the specified covariances not only shrinks (compare the width of the histograms for $d_x=2$ and $d_x=2000$), but also includes smaller values. 

Following the above experiment, we initialize $\beta_\phi$ at $10^{-2}$ in all of our studies in this paper since the input function is typically discretized at a large number of points. This value is obviously not an optimal choice so in the case of zero-mean GPs we associate each input feature with its own $\beta$ and then fine tune them via MLE. In the case of NN-mean GPs, we do not optimize these initial values (due to computational costs and numerical instabilities) and rely on the neural operator part of the GP to correct (if needed) how the correlations between $\phi(u)$ and $\phi(u')$ are quantified. 

Our rationale for initializing $\beta_\phi$ at a small value, \eg $10^{-2}$, can also be explained as follows. 
Consider the Gaussian kernel $c(u,u^\prime; \beta, \sigma^2=1)$:
\begin{equation}
    c(u,u^\prime; \beta, \sigma^2=1)=
    \exp \braces{- \beta \bars{\bars{u-u'}}^2},
    \label{eq gaussian distance functions}
\end{equation}
where $u(x)$ and $u^\prime(x)$ are one-dimensional functions and $\bars{\bars{\cdot}}$ is the $L2$ norm that can be approximated as:
\begin{equation}
    \bars{\bars{u-u'}}^2 = \int_\Omega (u(x)-u^\prime(x))^2dx \approx \sum_{i=1}^p (u(x_i)-u^\prime(x_i))^2\Delta x_i,
    \label{eq l2 norm functions}
\end{equation}
where $\Delta x_i$ is the discretization step. Assuming the points are equally spaced, \Cref{eq gaussian distance functions} simplifies to:
\begin{equation}
    c(u,u^\prime; \beta, \sigma^2=1)=
    \exp \braces{- \underbrace{\beta \Delta x}_{\beta_u} \sum_{i=1}^p (u(x_i)-u^\prime(x_i))^2}.
    \label{eq gaussian distance functions simplified}
\end{equation}
We observe that for function inputs the effective length-scale parameter $\beta_u$ corresponds to the product of the length-scale $\beta$ and the discretization step $\Delta x$, which is generally small. So, compared to the case where the features are scalars by nature, smaller values should be chosen for the effective length-scale parameter if the features are obtained by finely discretizing a function.

\begin{figure*}[!t]
    \centering
    \begin{subfigure}[t]{0.8\textwidth}
        \centering
        \includegraphics[width=1.00\columnwidth]{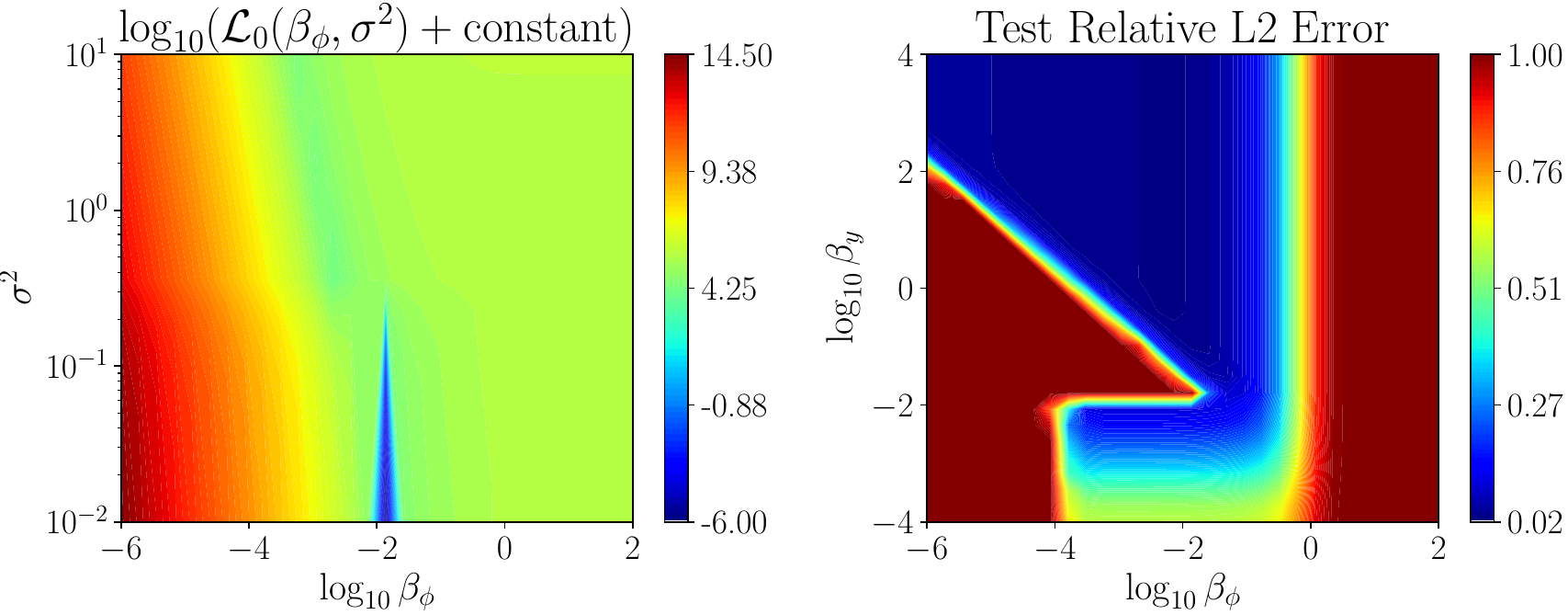}
        \captionsetup{justification=centering}
        \caption{Loss and error profiles for the $1D$ Burgers' problem in \Cref{eq Burgers main}.}
        \label{fig burgers profiles}
    \end{subfigure}
    \\
    \begin{subfigure}[t]{0.8\textwidth}
        \centering
        \includegraphics[width=1.00\columnwidth]{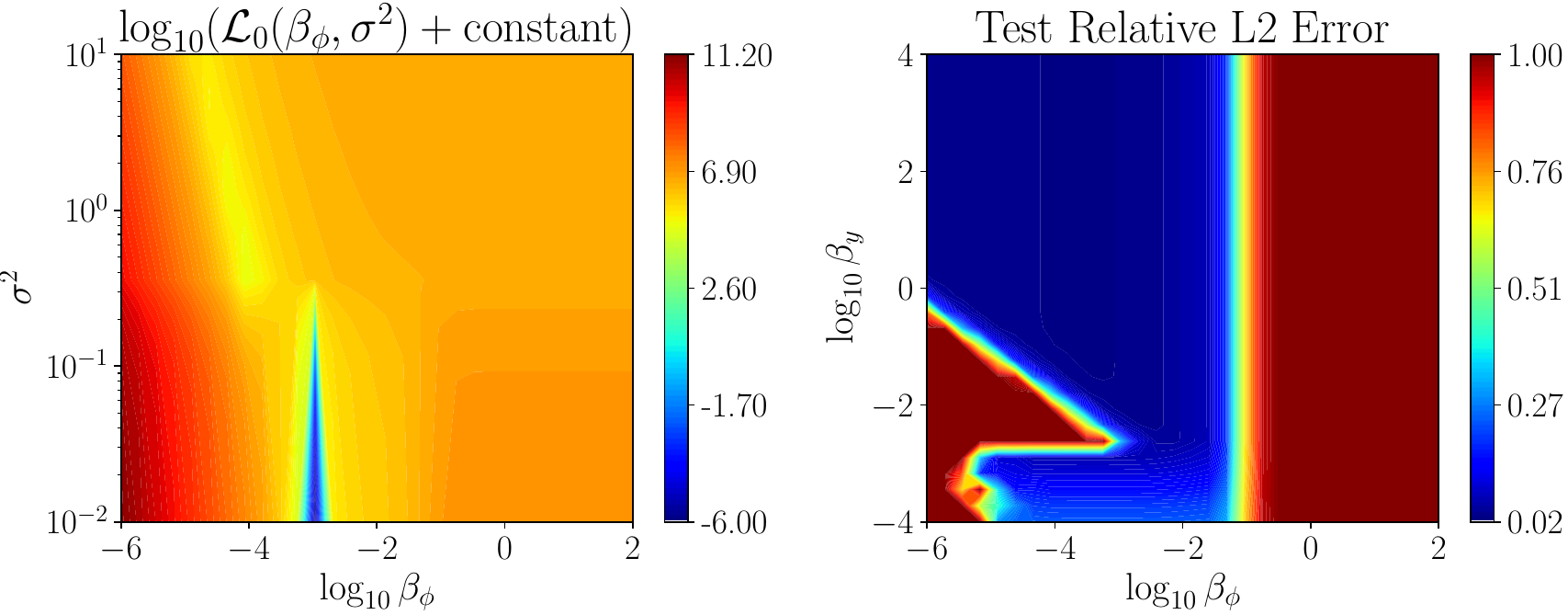}
        \captionsetup{justification=centering}
        \caption{Loss and error profiles for the Darcy problem in \Cref{eq Darcy main}.}
        \label{fig darcy profiles}
    \end{subfigure}%
    \caption{\textbf{Loss and error profiles as functions of kernel parameters in zero-mean GPs:} A wide range of parameter combinations provide optimal loss and error values, indicating that proper initialization is relatively easy and results in good performance even without training. $\beta_y = 10^{3}$ is used for the two loss profiles on the left-hand side of \textbf{(a)} and \textbf{(b)}. Note that we do not show the dependency of the test relative $L2$ error with respect to $\sigma^2_\phi$ since this does not affect \Cref{eq expected value posterior one sample}. For Darcy, $\beta_y$ has two components and so for plotting the error map we presume that these components are equal (a similar assumption is made for $\beta_\phi$).}
    \label{fig profiles}
\end{figure*}

Once again, these findings are also validated in \Cref{fig profiles}(a.2) and \Cref{fig profiles}(b.2), where we plot the test relative $L2$ errors as functions of $\beta_y$ and $\beta_\phi$ for the Burgers' and Darcy benchmarks. Our experiments suggest that the range of values for $\beta_\phi$ (corresponding to the functional input) that minimize the test error is comparably smaller to $\beta_y$ (corresponding to scalar inputs). An interesting observation from \Cref{fig profiles}(a.2) and \Cref{fig profiles}(b.2) is that the test error is minimized across a broad range of $\beta_y$ and $\beta_\phi$ values. This suggests that a zero-mean GP, when properly initialized, can achieve high accuracy without requiring any training.

To examine the interaction between the initialization of $\beta_y$ and $\beta_\phi$, we consider the Burgers' and Darcy problems studied in \Cref{sec results}. Applying zero-mean GPs to these two problems, we visualize the training loss and test error maps as a function of the kernel parameters in \Cref{fig profiles}. We observe that in all cases our suggested initial values provide minimal values for these metrics in both problems. Interestingly, we notice that many parameter combinations effectively minimize the loss and error maps in both problems, indicating that a high accuracy is obtained by a \textit{zero-shot} zero-mean GP whose kernel parameters are properly initialized but never optimized. We examine the performance of zero-shot and one-shot zero-mean GPs in \Cref{sec results} which benefit from zero and one optimization iterations, respectively.

\subsubsection{Parameter Optimization in NN-mean GPs} \label{subsubsec numerical instability}

Non-zero mean functions have the potential to increase the extrapolation capabilities of GPs \cite{iwata2017improving} and they have been effectively employed for materials modeling \cite{RN1558} or to solve forward and inverse problems in PDEs \cite{mora2024neural}. Specifically, parameterizing the mean function of a GP by a deep NN provides the means to leverage the strengths of both kernels and NNs as the former effectively learns local trends by interpolating neighboring points in regions with abundant data, while the latter improves generalization by learning complex distributed representations for unseen inputs. However, this added model complexity increases the risk of overfitting and can lead to numerical instability during optimization \cite{iwata2017improving}. We demonstrate these issues with a simple regression example which motivates our decision for fixing the kernel parameters while optimizing $\theta$. 

\begin{figure}[!h]
    \centering
        \includegraphics[width=1.0\linewidth]{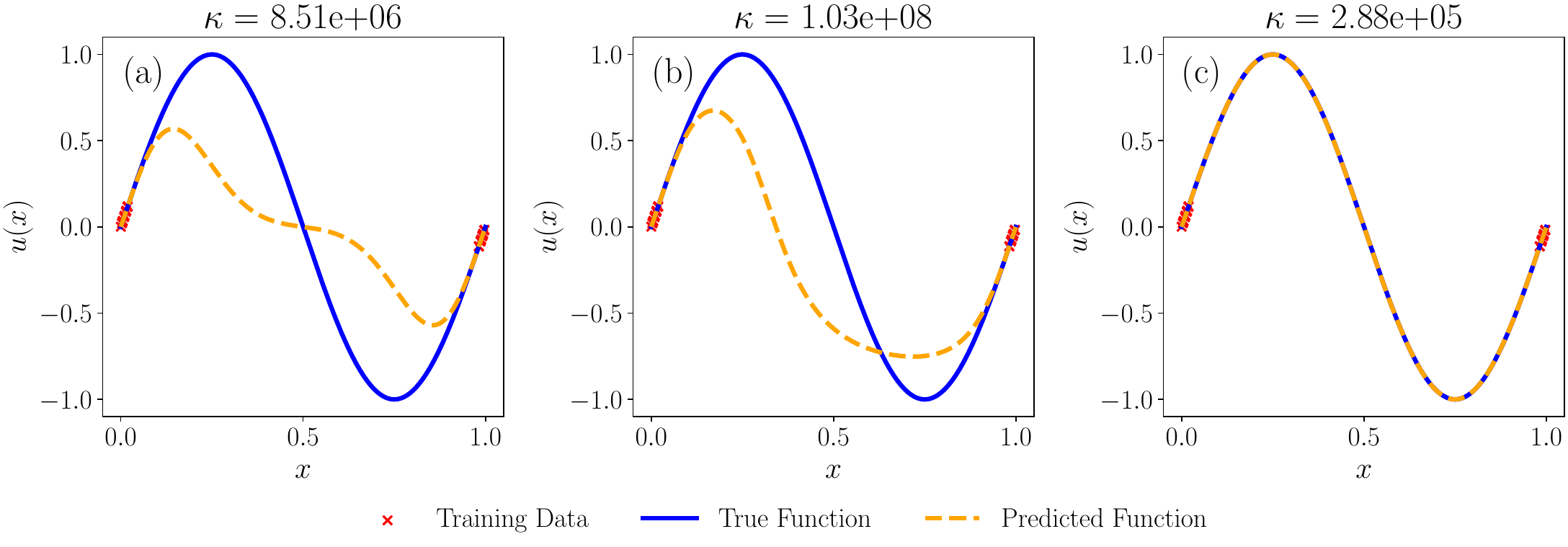}
    \caption{\textbf{Parameter optimization in GPs:} We consider three scenarios with a zero-mean GP in \textbf{(a)} and NN-mean GPs in \textbf{(b)} and \textbf{(c)}. The differences in training settings across these three cases are detailed in the text.}
    \label{fig joint opt}
\end{figure}

We again consider the function $u(x) = \sin(2\pi x)$, $x \in [0, 1]$ and sample $N=20$ points from it such that the samples are concentrated close to the lower- and upper-bounds of $x$, see the red crosses in \Cref{fig joint opt}. We fit three GPs to this dataset:
\begin{itemize}
    \item Zero-mean GP: We set the kernel to $c(x, x'; \beta, \sigma^2 = 1)$ and use MLE to optimize $\beta$.
    \item NN-mean GP: We set the kernel and the mean function to, respectively, $c(x, x'; \beta, \sigma^2 = 1)$ and a fully-connected feed-forward NN with $4$ layers of $20$ neurons. We use MLE to jointly optimize the parameters of the kernel and the NN. 
    \item Physics-informed NN-mean GP: We set the kernel to $c(x, x'; \beta = 10^3, \sigma^2 = 1)$ and optimize the NN parameters to minimize a weighted combination of MLE and MSE where the latter encourages the GP-based predictor to satisfy the ordinary differential equation $\frac{du}{dx} = 2 \pi \cos{(2 \pi x)}$.
\end{itemize}

The results of our studies are summarized in \Cref{fig joint opt} and demonstrate that the zero-mean GP (left plot) expectedly fails to generalize to regions where there are no training samples. This issue is slightly alleviated in the case of the NN-mean GP (middle plot) at the expense of worsening the condition number, $\kappa$, of the covariance matrix. The increase in $\kappa$ is due to the fact that the NN mean has sufficient parameters to interpolate the data which causes the kernel to model very small residuals and, in turn, be numerically ill-conditioned. The magnitude of $\kappa$ is directly related to the number of parameters of the mean function which is typically a very large number in neural operators. 

The results of the third scenario are provided in \Cref{fig joint opt}(c) where we observe that the physics of the problem is effectively used to learn $u(x)$ with high accuracy without increasing the condition number of the covariance matrix. The training mechanism corresponding to this scenario is similar to the one explained in \Cref{subsec physics operator learning} for operator learning.

\subsection{Comparison to Optimal Recovery-based Operator Learning} \label{subsec comparison}
The optimal recovery method of \cite{batlle2024kernel} approximates the operator $\mathGdagger$ via the following explicit representer formula:
\begin{equation}
    \mathcal{\bar{G}}(u^*)(y) = c_\nu(y, \Yb) c_\nu(\Yb, \Yb)^{-1} \Lambda(\phi(u^*),\Ub) \Lambda(\Ub,\Ub)^{-1} \Vb
    \label{eq operator houman 1}   
\end{equation}
where $\Lambda(\Ub,\Ub')$ is a matrix-valued kernel in the space of $m \times m$ matrices, $c_\nu(\cdot, \cdot)$ is the kernel associated with the RKHS of functions in $\mathV$, and $y$ denotes a location in the output space where observations are made. 
\cite{batlle2024kernel} assumes a diagonal kernel of the form $\Lambda(\Ub,\Ub') = c_\lambda(\Ub,\Ub')I$ where $I$ is the $m \times m $ identity matrix and $c_\lambda: \real^n \times \real^n \rightarrow \real$ is a real-valued kernel. Following these assumptions and using the Kronecker product, we can rewrite \Cref{eq operator houman 1} as:
\begin{equation}
    \mathcal{\bar{G}}(u^*)(y) = c_\nu(y, \Yb) c_\nu(\Yb, \Yb)^{-1} \brackets{c_\lambda\parens{\phi(u^*),\Ub} \otimes I } \brackets{c_\lambda(\Ub,\Ub) \otimes I}^{-1} \Vb.
    \label{eq operator houman}   
\end{equation}
To directly compare \Cref{eq operator houman} with our approach, we drop the mean function in our GP predictor to obtain:
\begin{equation}
    \E[\mathGtildedaggerp(\phi(u^*),\delta_{y^*}) | \mathD] =
    \brackets{c_{\phi}(\phi(u^*), \Ub) \otimes c_y(\Ystarb, \Yb)} \brackets{c_{\phi}(\Ub, \Ub) \otimes c_y(\Yb, \Yb)}^{-1} \Vb.
    \label{eq operator ours}   
\end{equation}
Comparing \Cref{eq operator houman,eq operator ours} highlights the main differences between the two approaches:
\begin{itemize}
    \item Unlike the optimal recovery method of \cite{batlle2024kernel}, we account for the correlations in the output function space through $c_y(\cdot, \cdot)$. Hence, we solve a single regression problem as opposed to solving $m$ independent optimal recovery ones at the expense of $(1)$ using all the data at once, and $(2)$ having far fewer hyperparameters (and hence flexibility).
    \item \Cref{eq operator houman} directly approximates $\mathGdagger$ which explains the presence of the operator-valued kernels. Instead, our method approximates $\mathGdagger(u^*)(y^*)$, \ie the \textit{evaluation} of the output function at the \textit{arbitrary} query point $y^*$ for the given input function $u^*$, see \Cref{fig diagram}. This feature enables us to incorporate additional physical constraints into the approximation process as detailed in \Cref{subsec physics operator learning}.    
\end{itemize}
We also note that our approach can also use deep NNs as the mean function which has the advantage of reducing the sole reliance of the predictor on the kernel whose misspecification can substantially reduce the performance of zero-mean GPs. 




\subsection{Inference Complexity} \label{subsec complexity}
We quantify the evaluation costs of our predictor by computing the floating point operators (FLOPs) per a test sample \cite{de2022cost}.
We assume that the product between two matrices $\Ab \in \real^{n\times m}$ and $\boldsymbol{B} \in \real^{m \times q}$ has a cost of $nq(2m-1)$ flops, vectorization has no costs, and that a nonlinear function acting upon an input vector of size $n$ (e.g., activation function in NNs or kernel evaluation in GPs) costs $n$ flops. Additionally, we assume the cost of constructing the $m \times n$ covariance matrix between two matrices $\Ab \in \real^{m\times q}$ and $\boldsymbol{B} \in \real^{n \times q}$ using the Gaussian kernel in \Cref{eq gaussian kernel} is $mn(4q + 2)$.

We observe in \Cref{eq expected value posterior one sample} that the $m \times N$ matrix $\CYb^{-1}\parens{\Vb - \Mb} \Cphib^{-1}$ can be stored from training and therefore we do not attribute any associated cost to that operation at the inference stage. Hence, the cost of evaluating the learned operator in our framework consists of 
$(1)$ evaluation cost of the mean function which is either zero for a zero-mean GP, or $\mathC_m$ for a non-zero mean GP, 
$(2)$ building $c_y(\Ystarb, \Yb; \widehat{\beta}_y) \in \real^{m \times m}$ and $c_{\phi}(\phi(u^*),  \phi(\Ub); \widehat{\beta}_\phi, \widehat \sigma^2_\phi) \in \real^{N \times 1}$, which have costs of $m^2(4d + 2)$ and $N(4n + 2)$, respectively, and 
$(3)$ matrix multiplication costs in $\vecOp{\parens{c_y(\Ystarb, \Yb; \widehat{\beta}_y) \CYb^{-1}\parens{\Vb^s - \Mb^s} \Cphib^{-1} c_{\phi}(\phi(u^*),  \Ub; \widehat{\beta}_\phi, \widehat \sigma^2_\phi) }}$ which sum up to $m(2N - 1) + m(2m - 1)$.
Given these cost components, the total inference cost of predicting the output function at $m$ points for a given test input function discretized at $n$ points is:
\begin{equation}
    \begin{aligned}
        \mathC &= \mathC_m + m^2(4d + 2) + N(4n + 2) + m(2N - 1) + m(2m - 1) \\
        & = \mathC_m + 4m^2(d + 1) - 2m + 2N(2n + 2m + 1),
    \end{aligned}
    \label{eq evaluation cost}
\end{equation}
where $N$ is the number of training samples and $\mathC_m$ can be calculated as detailed in \cite{de2022cost} if neural operators such as DeepONet or FNO are used as the mean function in our approach.

\Cref{eq evaluation cost} indicates the price that one has to pay for embedding a neural operator in our framework. This additional cost scales linearly in $N$ but incurring it, as argued in \Cref{sec results} based on our comparative studies, is justified since GPs with neural operators as their mean function provide higher accuracy than the base neural operators especially in extrapolation. 



    \section{Numerical Studies and Discussions} \label{sec results}
We evaluate the performance of our approach in data-driven and physics-informed operator learning problems in \Cref{subsec results data driven scalar,subsec results data driven vector,subsec results physics-informed} where we consider both single- and multi-output operators. While we can leverage any neural operator as the mean function in our approach, herein we only consider FNO \cite{li2020fourier} and DeepONet \cite{lu2021learning} which are two of the most popular neural operators. Throughout, we use the relative $L2$ metric defined below to quantify the accuracy on test data:
\begin{equation}
    L2 = \frac{1}{N_\text{test}} \sum_{i=1}^{N_\text{test}} \frac{\left\|\mathGdagger(u^i)-\mathGtildedaggerp(u^i)\right\|_{\mathV}}{\left\|\mathGdagger(u^i)\right\|_{\mathV}}
    \label{eq l2 definition}
\end{equation}
where $N_\text{test}$ is the number of test samples, $\mathGdagger$ is the target operator and $\mathGtildedaggerp$ is the predicted operator. We conclude our studies in \Cref{subsec training time} by commenting on the training cost of different approaches. 

\subsection{Data-driven Operator Learning: Single-output Operators} \label{subsec results data driven scalar}

We consider four canonical examples that are routinely studied in the relevant literature \cite{batlle2024kernel, de2022cost, lu2022comprehensive, li2020fourier}. These examples are defined below and used to compare our approach to DeepONet \cite{lu2021learning}, FNO \cite{li2020fourier}, and GP-OR which is the GP-based optimal recovery approach of \cite{batlle2024kernel}. 
We take the training data from \cite{lu2022comprehensive} and use the reported accuracy metrics in \cite{batlle2024kernel} for these three baselines. 
In our approach, we consider three cases: a zero-mean GP, DeepONet-mean GP, and FNO-mean GP. In the latter two cases, we set up the architecture of the neural operators to match those of the baselines\footnote{In some cases, insufficient information about the architecture is available so we highlight the potential differences when applicable.}. We also train our models with the same data used for training the baselines. 
We train our GP-based models via Adam for $3000$ training epochs and set the learning rate to $10^{-3}$ and $10^{-2}$ for NN-mean and zero-mean cases, respectively. 


\noindent \textbf{Burgers' Equation:}
We consider the $1D$ Burgers' equation with periodic BCs:
\begin{equation}
    \begin{aligned}
        &\ut + u \ux = \nu \uxx, && \forall (x,t) \in (0,1) \times (0, 1], \\
        &u(x, 0) = u_0(x), && \forall x \in (0, 1) \\
    \end{aligned}
    \label{eq Burgers main}
\end{equation}
with $\nu = 0.1$. The operator that we aim to learn is $\mathGdagger: u(x,0) \rightarrow u(x,1)$, \ie the mapping from the IC to the solution at time $t=1$. 
The dataset is originally generated in \cite{li2020fourier} where the IC is sampled from a Gaussian random field with a Riesz kernel. As in \cite{lu2022comprehensive}, we use $1000$ samples for training and $200$ for testing with a spatial resolution of $128$ points, i.e., $p=q=128$.

\noindent \textbf{Darcy Flow:}
The $2D$ Darcy flow problem with zero BCs is formulated as:
\begin{equation}
    \begin{aligned}
        &-\nabla \cdot (a(x,y) \nabla u(x,y)) = f(x,y), && \forall (x,y)\times (0,1)^2\\
        &u(x, y) = 0, && \forall (x,y) \in \partial(0, 1)^2 \\
    \end{aligned}
    \label{eq Darcy main}
\end{equation}
where $a(x,y)$ is the permeability field and $f(x,y)$ is a source term. For a fixed $f(x,y)$, we aim to learn $\mathGdagger: a(x,y) \rightarrow u(x,y)$, \ie the mapping from the permeability field to the solution in the entire domain. The dataset is extracted from \cite{lu2022comprehensive} and we use $1000$ samples for training and $200$ for testing where the input and output functions are available on a $29\times 29$ grid. The coefficient field $a(x,y)$ is obtained via $a=\xi(\mu)$ where $\mu \sim \mathcal{GP}(0,(-\Delta+9I)^{-2})$ is a Gaussian random field with zero Neumann BCs on the Laplacian and $\xi$ is a function that maps positive values to $12$ and negative values to $3$ \cite{lu2022comprehensive}.

\noindent \textbf{Advection Equation:}
We consider the $1D$ advection equation with periodic BCs:
\begin{equation}
    \begin{aligned}
        &\ut + \ux = 0, && \forall (x,t) \in (0,1) \times (0, 1], \\
        &u(x, 0) = u_0(x), && \forall x \in (0, 1) \\
    \end{aligned}
    \label{eq Advection main}
\end{equation}
where $\nu = 0.1$ and our goal is to approximate $\mathGdagger: u(x,0) \rightarrow u(x,0.5)$, \ie the mapping from the IC to the solution at time $t=0.5$. 
The dataset is originally generated by \cite{lu2022comprehensive} where the IC is designed as a square wave centered at $x=c$ with width $w$ and height $h$, \ie $u_0(x)=h\textbf{1}_{\braces{c-\frac{w}{2}},\braces{c+\frac{w}{2}}}$. The parameters $c,w$ and $h$ are randomly drawn from the uniform distribution $U([0.3, 0.7]\times [0.3, 0.6] \times [1,2])$. We use $1000$ samples for training and $200$ for testing whose input and output functions are discretized via $40$ equally-spaced grid points.

\noindent \textbf{Structural Mechanics:}
We consider the equation governing the displacement field $u$ of an elastic solid under small deformations:
\begin{equation}
    \begin{aligned}
        & \nabla \cdot \sigma = 0, && \text{in } D, \\
        &u = \bar{u}, && \text{on } \Gamma_u, \\
        &\nabla \cdot n = \bar{t}, && \text{on } \Gamma_t, \\
    \end{aligned}
    \label{eq Structural main}
\end{equation}
where $D=(0,1)^2$ and the boundary $\partial D$ is divided into $\Gamma_t = [0,1] \times {1}$ and $\Gamma_u = \partial D \setminus \Gamma_t$, \ie $\Gamma_u$ is the compliment of $\Gamma_t$. We aim to learn the operator $\mathGdagger: \bar{t} \rightarrow \sigma$, \ie the mapping between the $1D$ load $\bar{t}$ applied on $\Gamma_t$ to the von Misses stress field $\sigma$. The dataset is extracted from \cite{de2022cost} where $\bar{t}$ is drawn from a Gaussian random field $\mathcal{GP}(100,400^2(-\nabla + 3^2 I)^{-1})$ where $\nabla$ is the Laplacian with Neumann BCs applied to the space of functions with a spatial mean of zero. The output function $\sigma$ is obtained via the FEM. We use $1250$ samples for training and $20{,}000$ for testing. The input function is sampled on $41$ discretized points and the output function is interpolated on a regular $41\times 41$ grid.

The results of our comparative studies are summarized in \Cref{tab results} and indicate that our approach provides very competitive results. We observe that except for the Darcy example with DeepONet\footnote{This may be due to the fact that we could not exactly replicate the training settings and model architecture since the corresponding reference \cite{lu2022comprehensive} lacks sufficient details.}, the performance of FNO and DeepONet consistently improves if they are used as the mean function of a GP. We attribute this desirable property to two features. 
First, due to the second term on the right-hand side of \Cref{eq expected value posterior on train samples} our predictor naturally interpolates the data and hence (a) its mean function (i.e., the neural operator) can focus more on learning the sample-to-sample correlations that are \textit{not} captured by the kernel, see the loss function in \Cref{eq nll kronecker nn mean}, and (b) the irregular part of the target operator (that is hard to approximate with a smooth kernel) can be captured by the Neural Operator. 
Second, our use of kernels provides a natural regularization scheme that improves generalizability. To see this, we analyze \Cref{eq nll} within a regularization framework which is also used in contexts such as kernel ridge regression \cite{RN332}. The second term in \Cref{eq nll} measures the data fit while the first term is a regularizer whose magnitude can be calculated based on the eigenvalues of $\Cb$. 
Since we freeze the kernel parameters in our NN-mean GPs, these eigenvalues are essentially fixed which imposes a smoothness condition on the underlying function space which is controlled by the decay rate of the sequence of eigenvalues. 

\begin{table}[!b]
    \centering
    \resizebox{\textwidth}{!}{%
    \begin{tabular}{lc|cc|cc|cc|cc|}
    \cline{3-10}
    \multicolumn{2}{l|}{} & \multicolumn{2}{c|}{\textbf{\begin{tabular}[c]{@{}c@{}}Burgers'\\ $N = 1000, p = 128, q = 128$\end{tabular}}} & \multicolumn{2}{c|}{\textbf{\begin{tabular}[c]{@{}c@{}}Darcy\\ $N = 1000, p = 29^2, q = 29^2$\end{tabular}}} & \multicolumn{2}{c|}{\textbf{\begin{tabular}[c]{@{}c@{}}Advection\\ $N = 1000, p = 40, q = 40$\end{tabular}}} & \multicolumn{2}{c|}{\textbf{\begin{tabular}[c]{@{}c@{}}Structural Mechanics\\ $N = 1250, p = 41, q = 41^2$\end{tabular}}} \\ \cline{3-10} 
     & \multicolumn{1}{l|}{} & \multicolumn{1}{c|}{\textbf{Error}} & \textbf{\# of Parameters} & \multicolumn{1}{c|}{\textbf{Error}} & \textbf{\# of Parameters} & \multicolumn{1}{c|}{\textbf{Error}} & \textbf{\# of Parameters} & \multicolumn{1}{c|}{\textbf{Error}} & \textbf{\# of Parameters} \\ \hline
    \multicolumn{2}{|c|}{\textbf{DeepONet}} & \multicolumn{1}{c|}{$2.15\%$} & $148{,}864$ & \multicolumn{1}{c|}{$2.91\%$} & $715{,}776$ & \multicolumn{1}{c|}{$0.22\%$} & $471{,}552$ & \multicolumn{1}{c|}{$8.70\%$} & $137{,}856$ \\ \hline
    \multicolumn{2}{|c|}{\textbf{FNO}} & \multicolumn{1}{c|}{$1.93\%$} & $320{,}705$ & \multicolumn{1}{c|}{$2.41\%$} & $6{,}304{,}257$ & \multicolumn{1}{c|}{$0.66\%$} & $320{,}705$ & \multicolumn{1}{c|}{$6.62\%$} & $888{,}001$ \\ \hline
    \multicolumn{2}{|c|}{\textbf{GP-OR}} & \multicolumn{1}{c|}{$2.15\%$} & $1280$ & \multicolumn{1}{c|}{$2.75\%$} & $169{,}882$ & \multicolumn{1}{c|}{\bm{$2.75\mathrm{e}{-3}\%$}} & $1{,}600$ & \multicolumn{1}{c|}{$6.95\%$} & $68{,}921$ \\ \hline
    \multicolumn{1}{|l|}{\multirow{5}{*}{\textbf{Our GP}}} & \textbf{0-shot Zero-mean} & \multicolumn{1}{c|}{$2.90\%$} & \bm{$129$} & \multicolumn{1}{c|}{$5.14\%$} & \bm{$201$} & \multicolumn{1}{c|}{$0.17\%$} & \bm{$41$} & \multicolumn{1}{c|}{$7.13\%$} & \bm{$42$} \\ \cline{2-10} 
    \multicolumn{1}{|l|}{} & \textbf{1-shot Zero-mean} & \multicolumn{1}{c|}{$2.90\%$} & \bm{$129$} & \multicolumn{1}{c|}{$5.01\%$} & \bm{$201$} & \multicolumn{1}{c|}{$0.17\%$} & \bm{$41$} & \multicolumn{1}{c|}{$7.13\%$} & \bm{$42$} \\ \cline{2-10} 
    \multicolumn{1}{|l|}{} & \textbf{Zero-mean} & \multicolumn{1}{c|}{$2.99\%$} & \bm{$129$} & \multicolumn{1}{c|}{$2.89\%$} & \bm{$201$} & \multicolumn{1}{c|}{$4.11\mathrm{e}{-3}\%$} & \bm{$41$} & \multicolumn{1}{c|}{$6.74\%$} & \bm{$42$} \\ \cline{2-10} 
    \multicolumn{1}{|l|}{} & \textbf{DeepONet-mean} & \multicolumn{1}{c|}{$1.84\%$} & $148{,}864$ & \multicolumn{1}{c|}{$3.44\%$} & $715{,}776$ & \multicolumn{1}{c|}{$0.18\%$} & $471{,}552$ & \multicolumn{1}{c|}{$7.12\%$} & $137{,}856$ \\ \cline{2-10} 
    \multicolumn{1}{|l|}{} & \textbf{FNO-mean} & \multicolumn{1}{c|}{\bm{$0.08\%$}} & $320{,}705$ & \multicolumn{1}{c|}{\bm{$2.19\%$}} & $6{,}304{,}257$ & \multicolumn{1}{c|}{$0.23\%$} & $320{,}705$ & \multicolumn{1}{c|}{\bm{$6.49\%$}} & $888{,}001$ \\ \hline
    \end{tabular}
    }
    \caption{\textbf{Summary results on four benchmark problems:} We report the $L2$ relative test error for four different operator learning methods. The errors in the first three rows are directly taken from the literature \cite{de2022cost, batlle2024kernel}. In our approach, we set the mean function to either $(1)$ zero in which case we also provide the error metrics without training and after one optimization iteration, or $(2)$ DeepONet and FNO whose architectures are exactly the same as those in rows $1$ and $2$. In the case of Darcy, we follow the references and apply PCA to the discretized input function and keep the first $200$ PCs as the features for the kernel part of our GP-based methods.}
    \label{tab results}
\end{table}

We also observe in \Cref{tab results} that our zero-mean GP is highly efficient in that it achieves small errors with orders of magnitude fewer hyperparameters. Due to the proper initialization of the kernel parameters, these GPs provide quite competitive results even prior to parameter optimization or just after one training epoch (referred to as $0-$shot and $1-$shot, respectively, in the table).

\begin{figure}[!h]
    \centering
        \includegraphics[width=1.0\linewidth]{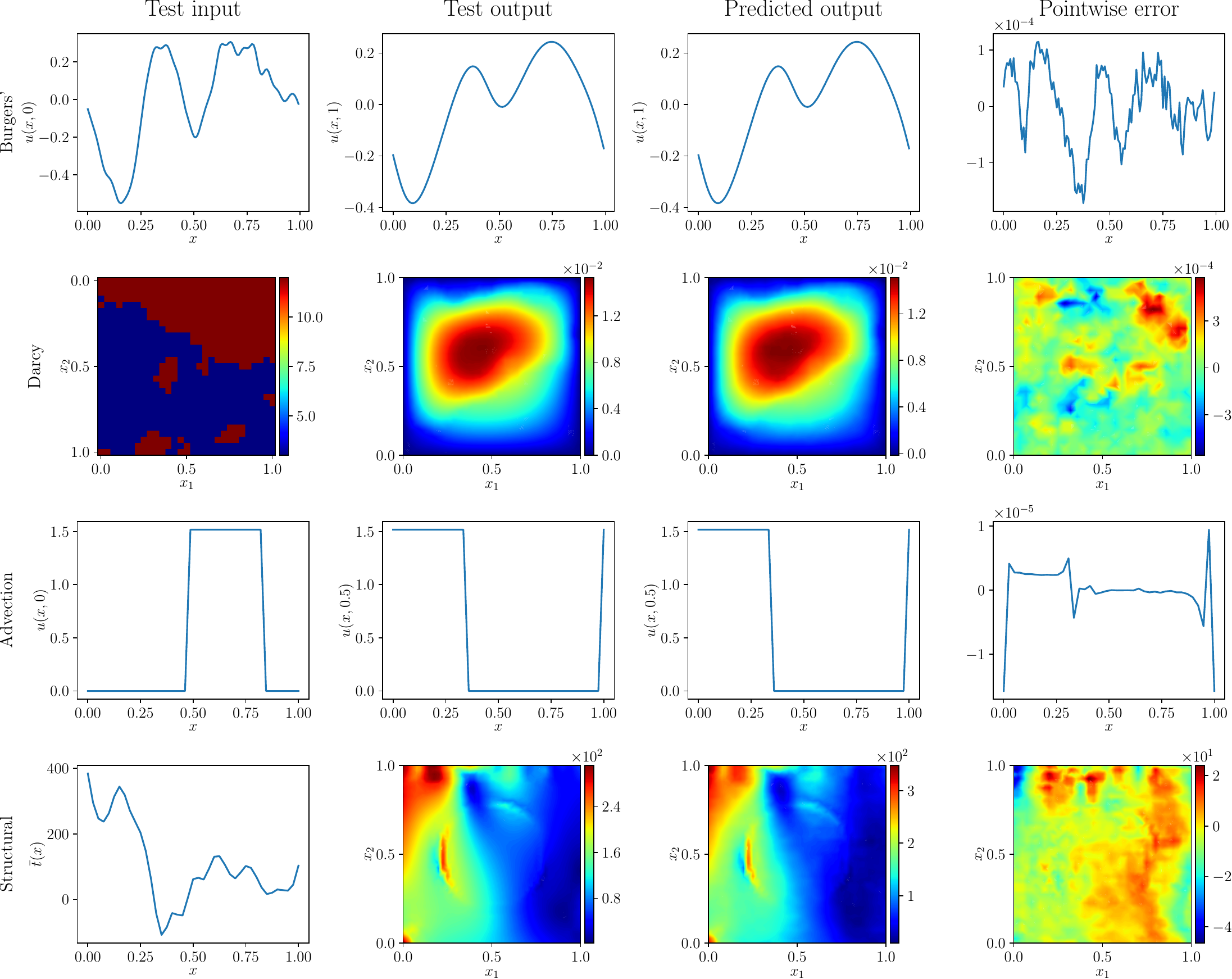}
    \caption{\textbf{Example test inputs, outputs, and predictions:} We provide one example from each problem where the predicted outputs correspond to the best version of our approach in \Cref{tab results}, i.e., a zero-mean GP in the case of Advection problem and an FNO-mean GP for the other three problems.}
    \label{fig results single output}
\end{figure}

In \Cref{fig results single output} we visualize example test inputs, outputs, and predictions with associated error maps. In these plots we use the best version of our approach, i.e., a zero-mean GP in the case of Advection problem and an FNO-mean GP for the other three cases. We observe that the majority of the errors are concentrated at regions with sharp gradients which correspond to the jumps in the Advection example and the large traction regions at the top left corner in the Structural Mechanics problem. 

\subsection{Data-driven Operator Learning: Multi-output Operators} \label{subsec results data driven vector}
\begin{figure*}[!t]
    \centering
    \begin{subfigure}[t]{0.48\textwidth}
        \centering
        \includegraphics[width=1.00\columnwidth]{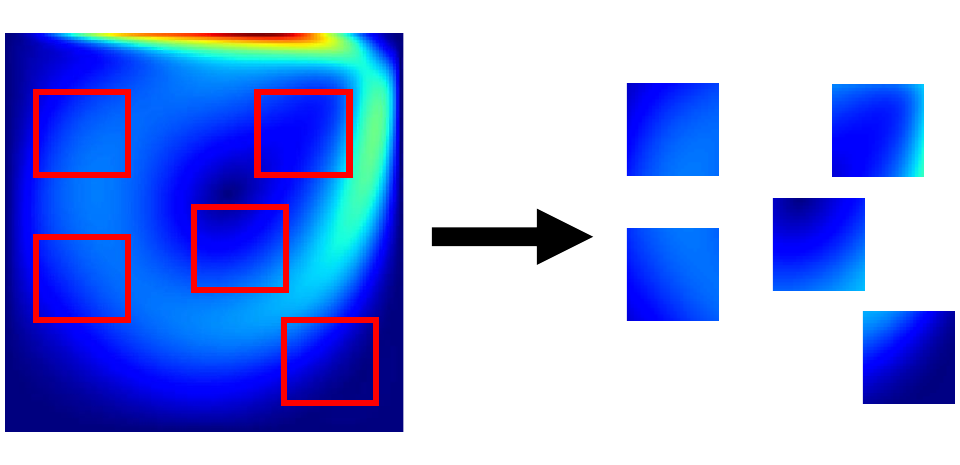}
        \captionsetup{justification=centering}
        \caption{Data collection.}
        \label{fig ldc sweeping}
    \end{subfigure}
    \begin{subfigure}[t]{0.48\textwidth}
        \centering
        \includegraphics[width=1.00\columnwidth]{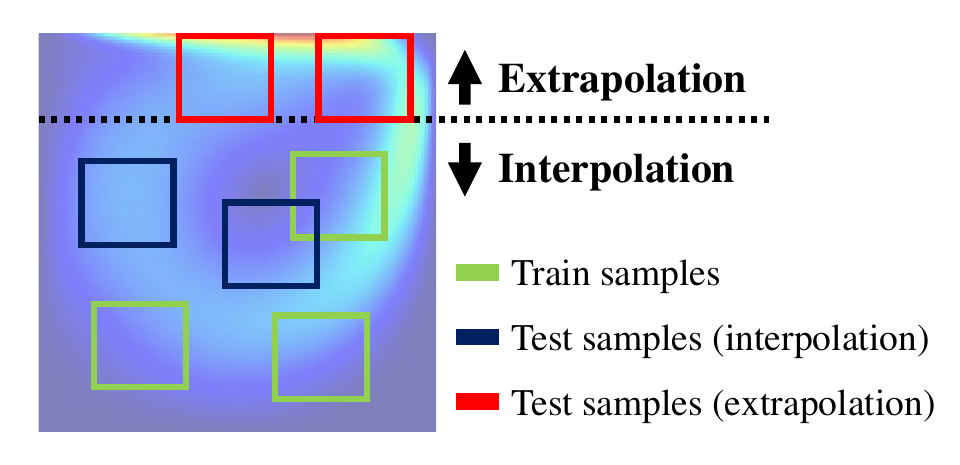}
        \captionsetup{justification=centering}
        \caption{Train vs test samples.}
        \label{fig ldc train vs test}
    \end{subfigure}%
    \caption{\textbf{Train and test data generation in the LDC problem:} \textbf{(a)} The $\brackets{0, 1}^2$ domain is swept with a smaller window to collect data. \textbf{(b)} If the center of the window is above the dashed line, the sample is used to assess the extrapolation error. Otherwise, the sample is used either for training or obtaining the interpolation error.}
    \label{fig ldc data collection}
\end{figure*}

To benchmark our approach for learning multi-output operators we use the $2D$ lid-driven cavity (LDC) problem which is governed by the incompressible Navier-Stokes equations:
\begin{equation}
    \begin{aligned}
        &\ux + \vy = 0, && \forall (x, y) \in (0,1)^2 \\
        &u \ux + v \uy = -\frac{1}{\rho}\px + \nu \parens{\uxx + \uyy}, && \forall (x, y) \in (0,1)^2 \\
        &u \vx + v \vy = -\frac{1}{\rho}\py + \nu \parens{\vxx + \vyy}, && \forall (x, y) \in (0,1)^2 \\
        &v(x, 0) = v(x, 1) = v(0, y) = v(1, y) = 0, && \forall x,y \in [0,1] \\
        &u(x, 0) = u(0, y) = u(1, y) = 0, && \forall x,y \in [0,1] \\
        &u(x, 1) = f(x), && \forall x \in [0, 1] \\      
        &p(0,0) = 0
    \end{aligned}    
    \label{eq ldc}
\end{equation}
where $\nu = 0.002$ is the kinematic viscosity, $\rho = 1.0$ denotes the density, and $f(x)$ is the lid velocity profile. To generate the data, we follow \cite{wang2022mosaic} and first solve \Cref{eq ldc} via the FEM for a variety of lid velocity profiles\footnote{The velocity profiles are taken from \cite{wang2022mosaic} who use GPs to generate them and solve \Cref{eq ldc} via OpenFOAM \cite{Jasak20071}.} and then sweep the flow field via a window of size $0.25 \times 0.25$ to collect samples, see \Cref{fig ldc data collection}. We divide the resulting data into two parts: if the window is above the horizontal dashed line in \Cref{fig ldc train vs test} the corresponding sample is used to assess the extrapolation test error. Otherwise, the sample is used for either training or obtaining the interpolation test error. The samples positioned on the top part of the domain provide good candidates for evaluating the extrapolation capabilities of the models since the solution fields inside them contain much larger velocity magnitudes and gradients than those in the bottom part of the domain.

The above procedure generates a total of $961$ samples with observations on a $30 \times 30$ regular grid in the $0.25 \times 0.25$ subdomain. The operator learning problems involves mapping the arbitrary set of subdomain BCs (pressure and velocity components) to the solution inside the $0.25 \times 0.25$ subdomain, \ie $\mathGdagger: [u_{\text{BC}},v_{\text{BC}},p_{\text{BC}}] \rightarrow [u,v,p]$. We use velocity magnitude to calculate the interpolation/extrapolation errors and train all models via Adam with $2000$ epochs and learning rates of $10^{-2}$ and $10^{-3}$ for the 0-mean and NN-mean GPs, respectively. All models with a DeepONet have $4-$ layer branch and trunk nets (each layer with $150$ neurons) and all FNO-based models have $8$ modes and $4$ channels. We use $500$ samples for training while $151$ and $31$ samples are used to obtain the interpolation and extrapolation errors, respectively. 

\begin{table}[!b]
    \centering
    \resizebox{1\textwidth}{!}{%
    \begin{tabular}{lc|c|cc|cc|}
    \cline{3-7}
    \multicolumn{2}{l|}{} & \multicolumn{1}{l|}{\multirow{2}{*}{\textbf{Number of Parameters}}} & \multicolumn{2}{c|}{\textbf{Interpolation Error}} & \multicolumn{2}{c|}{\textbf{Extrapolation Error}} \\ \cline{4-7} 
     & \multicolumn{1}{l|}{} & \multicolumn{1}{l|}{} & \multicolumn{1}{c|}{\textbf{Training via MLE}} & \textbf{Training via MSE} & \multicolumn{1}{c|}{\textbf{Training via MLE}} & \textbf{Training via MSE} \\ \hline
    \multicolumn{2}{|c|}{\textbf{DeepONet}} & $6{,}235$ & \multicolumn{1}{c|}{N/A} & $3.49\%$ & \multicolumn{1}{c|}{N/A} & $35.14\%$ \\ \hline
    \multicolumn{2}{|c|}{\textbf{FNO}} & $235{,}800$ & \multicolumn{1}{c|}{N/A} & $7.80\%$ & \multicolumn{1}{c|}{N/A} & $62.66\%$ \\ \hline
    \multicolumn{1}{|l|}{\multirow{5}{*}{\textbf{Ours}}} & \textbf{0-shot Zero-mean} & \bm{$361$} & \multicolumn{1}{c|}{$7.10\mathrm{e}{-3}\%$} & N/A & \multicolumn{1}{c|}{$24.48\%$} & N/A \\ \cline{2-7} 
    \multicolumn{1}{|l|}{} & \textbf{1-shot Zero-mean} & \bm{$361$} & \multicolumn{1}{c|}{$7.10\mathrm{e}{-3}\%$} & N/A & \multicolumn{1}{c|}{$24.48\%$} & N/A \\ \cline{2-7} 
    \multicolumn{1}{|l|}{} & \textbf{Zero-mean} & \bm{$361$} & \multicolumn{1}{c|}{$1.51\mathrm{e}{-2}\%$} & N/A & \multicolumn{1}{c|}{$28.76\%$} & N/A \\ \cline{2-7} 
    \multicolumn{1}{|l|}{} & \textbf{DeepONet-mean} & $6{,}235$ & \multicolumn{1}{c|}{\bm{$2.73\mathrm{e}{-3}\%$}} & \bm{$8.46\mathrm{e}{-3}\%$} & \multicolumn{1}{c|}{\bm{$22.77\%$}} & \bm{$23.48\%$} \\ \cline{2-7} 
    \multicolumn{1}{|l|}{} & \textbf{FNO-mean} & $235{,}800$ & \multicolumn{1}{c|}{$3.76\mathrm{e}{-3}\%$} & $9.70\mathrm{e}{-3}\%$ & \multicolumn{1}{c|}{$26.81\%$} & $34.60\%$ \\ \hline
    \end{tabular}
    }
    \caption{\textbf{Summary results for the LDC problem:} We use DeepONet and FNO for comparison in both interpolation and extrapolation tests. In our approach, we consider either these two neural operators as the mean function or use a zero-mean GP. In the former case, we train the GPs based on MLE and MSE to indicate the effect of the loss function on model performance. For the zero-mean GP, we also provide the errors prior to parameter optimization ($0-$shot) and after one training epoch ($1-$shot). The errors are reported for the velocity magnitude.}
    \label{tab results LDC}
\end{table}

The results of our studies are summarized in \Cref{tab results LDC} and indicate that our GP-based approach consistently outperforms FNO and DeepONet in this example. The best results are obtained via a GP which uses DeepONet as its mean function and is trained via MLE. Switching to MSE from MLE reduces the accuracy in both extrapolation and interpolation regardless of whether FNO or DeepONet is used as the mean function in our approach since MSE does not take the sample-to-sample correlations into account while MLE does, see \Cref{eq multitask nll kronecker non-zero-mean}.

Similar to the studies in \Cref{subsec results data driven scalar}, we observe in \Cref{tab results LDC} that our zero-mean GP provides very competitive results while having substantially fewer parameters. Interestingly, the performance of this GP deteriorates upon training as its $0-$shot and $1-$shot versions achieve smaller errors in both interpolation and extrapolation. This trend is due to the fact that the numbers in \Cref{tab results LDC} are based on velocity magnitude and do not show the positive effect of training on predicting the pressure field.

\begin{figure*}[!t]
    \centering
    \begin{subfigure}[t]{\textwidth}
        \centering
        \includegraphics[width=1.0\textwidth]{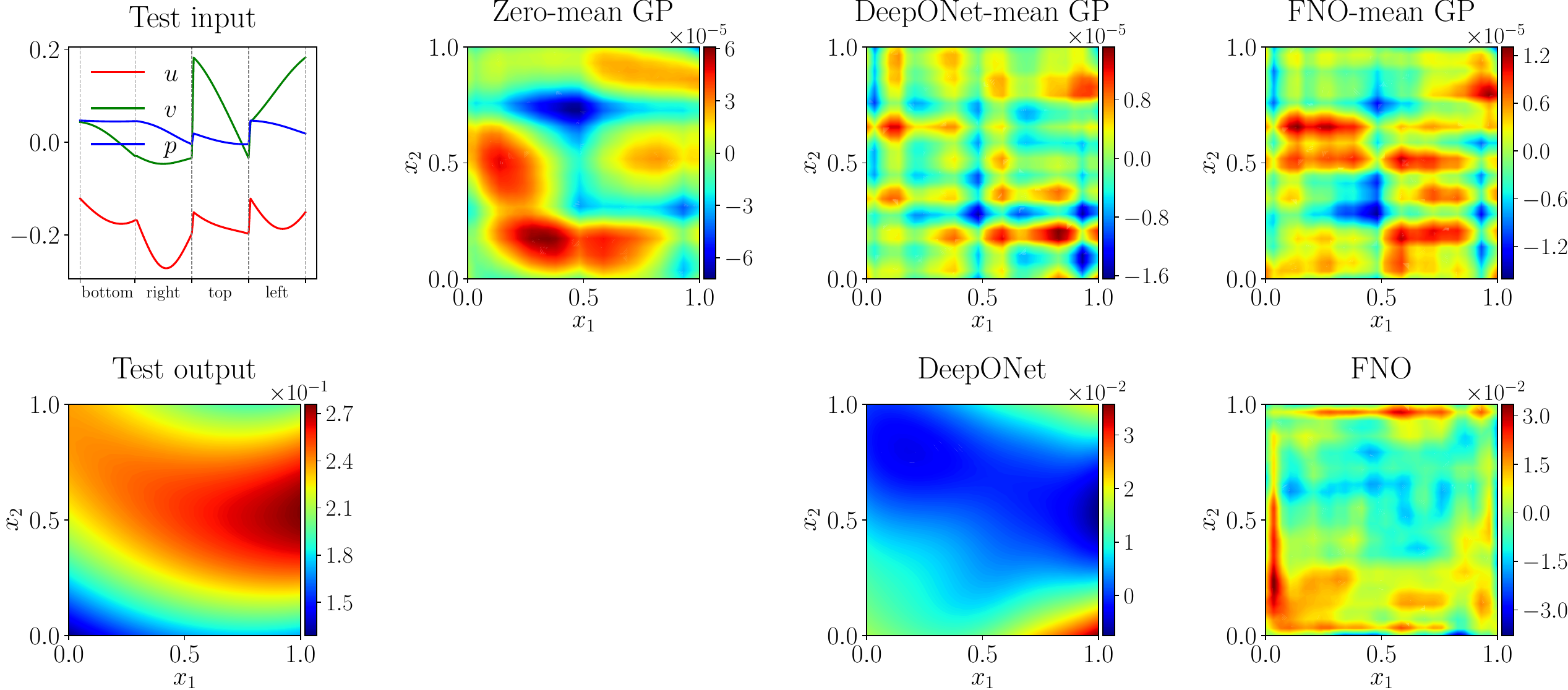}
        \captionsetup{justification=centering}
        \caption{Interpolation.}
        \label{fig ldc error maps interpolation}
    \end{subfigure}
    
    \vspace{\floatsep}  
    
    \begin{subfigure}[t]{\textwidth}
        \centering
        \includegraphics[width=1.0\textwidth]{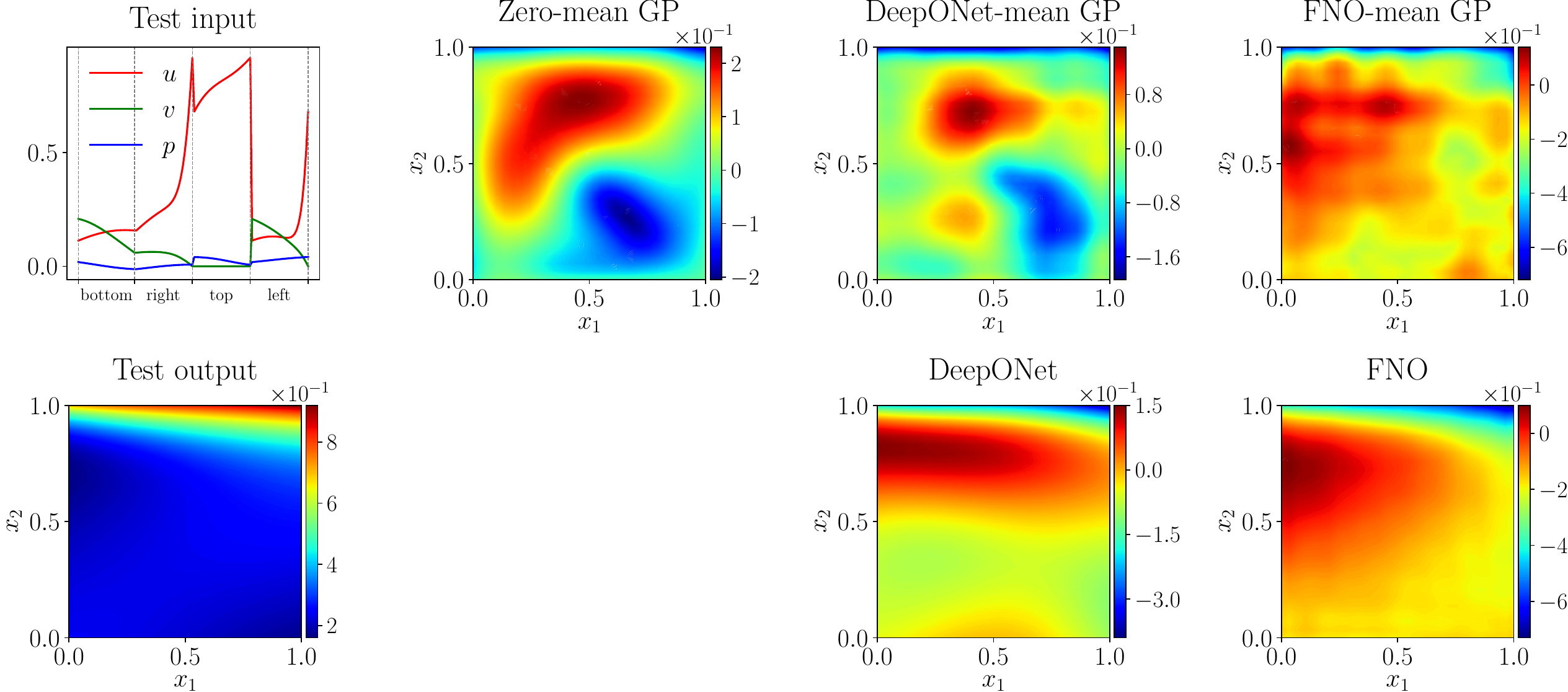}
        \captionsetup{justification=centering}
        \caption{Extrapolation.}
        \label{fig ldc error maps extrapolation}
    \end{subfigure}

    \caption{\textbf{Example input, output, and error maps on the velocity magnitudes in the LDC problem:} The test samples in extrapolation look substantially different than those in interpolation (compare the scales of the velocity magnitude profiles in the two test outputs). Our GP-based approach provides smaller errors in both scenarios especially in the interpolation case.}
    \label{fig ldc error maps}
\end{figure*}

To visually compare the accuracy of different methods in extrapolation and interpolation, in \Cref{fig ldc error maps} we provide the error maps corresponding to the velocity magnitude for two test cases. We observe in \Cref{fig ldc error maps interpolation} that all methods performs quite well but our GP-based approach achieves two orders of magnitude smaller errors compared to FNO and DeepONet. However, the performance of all methods expectedly decreases in \Cref{fig ldc error maps extrapolation} as the imposed BCs are substantially different than the ones that the models are trained on.

\subsection{Physics-informed Operator Learning} \label{subsec results physics-informed}
We now consider the effect of augmenting data with physics in two examples, the Burger's equation introduced in \Cref{subsec method summary} and the LDC problem in \Cref{subsubsec vectorvalued operators}. In the former example, we discretize the input function via $100$ points and record the output at a very coarse grid of size $12\times12$ and use $N_{pi} = 50, n_{PDE} = 100^2, n_{BC} = 100,$ and, $n_{IC} = 100$. In the LDC problem, each variable is discretized via $120$ points on the boundaries ($30$ points on each edge of the square subdomain in \Cref{fig ldc sweeping}) and the PDE solution is recorded at a fine mesh of size $30\times30$, and use $N_{pi} = 100, n_{PDE} = 30^2, n_{BC} = 30$.

We solve these two problems via FNO, DeepONet, and their physics-informed versions denoted by PI-DeepONet \cite{wang2021learning} and PINO \cite{li2024physics}. For our approach, we consider both zero-mean and NN-mean GPs where the latter is trained either with data or with both data and physics. In our NN-mean GP we only use DeepONet as we were unable to successfully integrate our code with PINO. All the models are trained via Adam with a learning rate of $10^{-3}$ which, similar to the previous sections, is increased to $10^{-2}$ for a zero-mean GP. We use $1000$ training epochs in all cases except for the physics-informed models of the LDC problem where the number of epochs is reduced to $500$ to decrease the computational costs. Since the density of the observations is very low in the Burgers' equation, we optimize $\beta_y$ along with $\beta_\phi$ and $\sigma^2$ via MLE.

\begin{figure}[!h]
    \centering
    \includegraphics[width=1.0\linewidth]{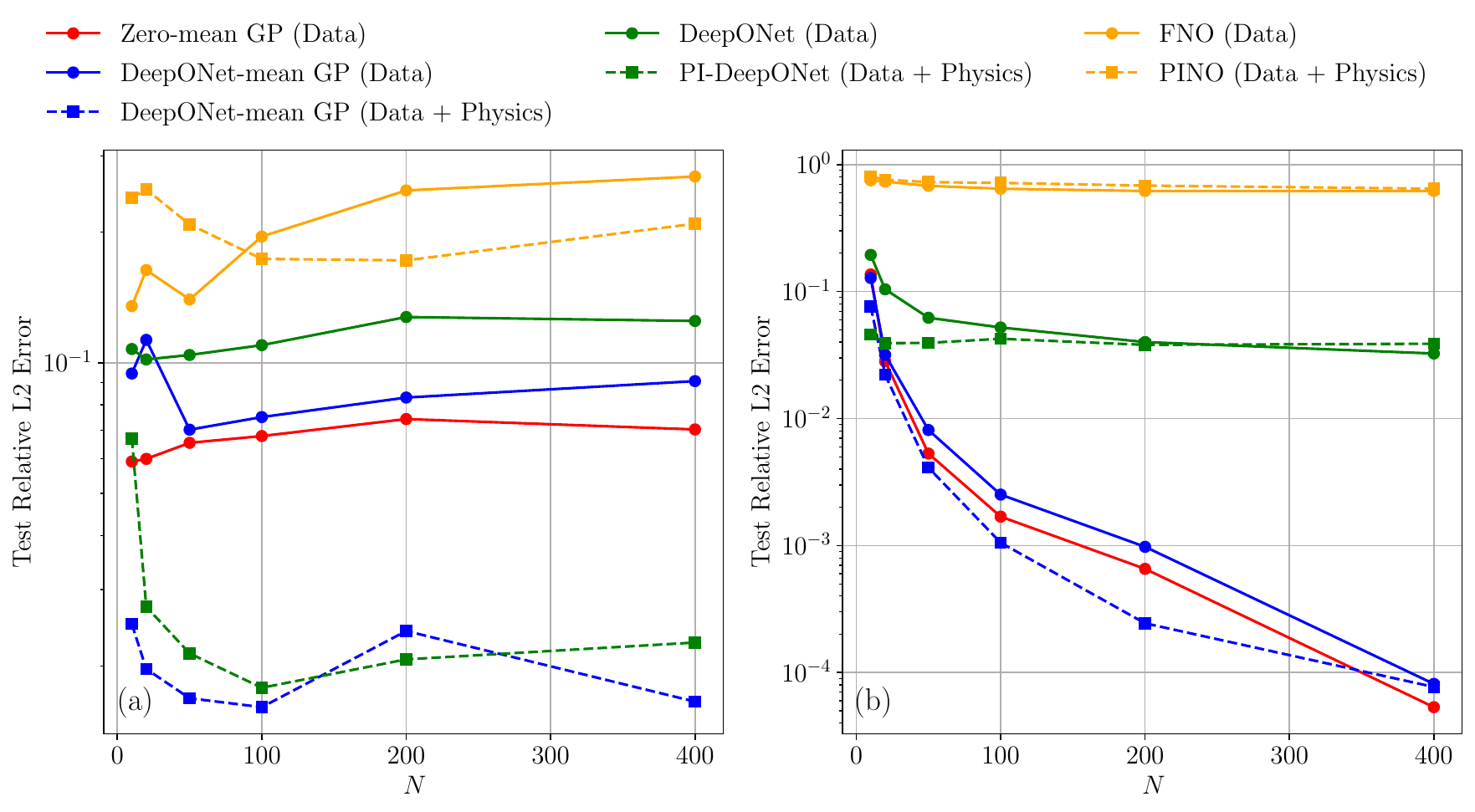}
    \caption{\textbf{Convergence studies:} We study the effect of data set size $N$ and augmenting the data with physics on the errors for various models in the case of Burgers' (left) and LDC (right) problems.}
    \label{fig results physics}
\end{figure}

The results of our studies are summarized in \Cref{fig results physics} where the errors on unseen data are reported for each model as $N$ increases. For the Burgers' equation where the observations are sparse, augmenting the training data with the physics substantially helps in the case of GP and DeepONet but not FNO (compare the dashed and solid lines with the same color). This difference is much smaller in the case of LDC where adding physics slightly improves GP and DeepONet when $N$ is very small. 

We notice as $N$ increases that the errors associated with the data-only models in the left panel of \Cref{fig results physics} slightly fluctuate while those in the right panel either decrease (in our approach) or negligibly change (FNO and DeepONet). The former trend is due to the fact that we observe the PDE solution at a very coarse grid and that the architecture and training mechanisms are not altered as $N$ increases. We attribute the latter trend to FNO and DeepONet's inability to effectively use data. However, once the exact same DeepONet is used as the mean function of a GP, the prediction accuracy dramatically improves (compare the solid green and blue lines in the right panel). See \Cref{subsec results data driven scalar} for our reasoning behind this trend. 

\subsection{Training Cost} \label{subsec training time}
We compare the training costs of our approach to those of FNO and DeepONet in \Cref{tab training time} where the costs are reported for one epoch of parameter optimization. We observe that the training cost of each approach varies depending on the dimensionality and resolution of the input/output functions in each problem. For instance, the zero-mean GP appears to have the highest training time per epoch for the Burgers' and Advection problems where the input and output functions are one-dimensional. However, it has the lowest training cost for the Structural Mechanics problem where the input and output functions are one- and two-dimensional, respectively. Incorporating neural operators within our GP framework increases the training time per epoch with respect to their baseline implementations, due to the additional computations involved in the loss function (compare \Cref{eq nll kronecker nn mean} to MSE). We believe this additional cost is justified since the non-zero mean GPs achieve higher accuracy compared to the standalone neural operators, as shown in \Cref{tab results} and \ref{tab results LDC}.  


\begin{table}[!h]
    \centering
    \resizebox{\textwidth}{!}{%
    \begin{tabular}{cc|c|c|c|c|c|}
        \cline{3-7}
        \multicolumn{2}{l|}{} & \textbf{Burgers'} & \textbf{Darcy} & \textbf{Advection} & \textbf{Structural Mechanics} & \textbf{Lid-driven Cavity} \\ \hline
        \multicolumn{2}{|c|}{\textbf{DeepONet}} & $0.04$ & $0.17$ & $0.03$ & $12.7$ & $0.22$ \\ \hline
        \multicolumn{2}{|c|}{\textbf{FNO}} & $0.06$ & $0.53$ & $0.03$ & $3.53$ & $0.03$ \\ \hline 
        \multicolumn{1}{|l|}{\multirow{3}{*}{\textbf{Ours}}} & \textbf{Zero-mean} & $0.12$ & $0.24$ & $0.12$ & $1.74$ & $0.20$ \\ \cline{2-7} 
        \multicolumn{1}{|l|}{} & \textbf{DeepONet-mean} & $0.06$ & $0.25$ & $0.05$ & $13.4$ & $0.27$ \\ \cline{2-7} 
        \multicolumn{1}{|l|}{} & \textbf{FNO-mean} & $0.07$ & $1.17$
         & $0.05$ & $3.66$ & $0.04$ \\ \hline
    \end{tabular}%
    }
    \caption{\textbf{Training time in seconds per epoch:} We report the costs for DeepONet, FNO, and our method with three different mean functions. The training time of each approach varies across the different problems based on the dimensionality of the input/output functions and their respective resolutions. For example, the zero-mean GP appears to be the slowest in problems where the input and output functions are one-dimensional, such as in the Burgers' or Advection equations, but becomes the fastest in the Structural Mechanics problem, where the input and output functions are one- and two-dimensional, respectively. Incorporating neural operators within our GP framework increases the training time per epoch with respect to their vanilla implementations due to the additional computations involved in the loss function (see \Cref{eq nll kronecker nn mean} versus MSE). However, this results in higher overall accuracy, as shown in \Cref{tab results} and \ref{tab results LDC}.} 
    
    \label{tab training time}
\end{table}

    \section{Conclusions} \label{sec conclusions}
We introduce a framework based on GPs for approximating single- or multi-output operators in either a purely data-driven context, or by using both data and physics. In the former case, we use MLE for parameter optimization and set the mean function of the GP to be either zero or a neural operator. In the physics-informed case, we employ GPs with neural operator mean functions whose parameters are estimated via a weighted combination of MLE and PDE residuals. 

In our framework, we devise a few strategies to ensure scalability to large data and high dimensions while minimizing the computational costs and numerical issues associated with inverting the GP's covariance matrix. Specifically, we $(1)$ leverage the data structure to formulate the GP's kernel based on the Kronecker product, and $(2)$ initialize the kernel parameters such that they need little to no tuning. In data-driven applications, these strategies enable us to build zero-mean GPs that provide high approximation accuracy in both single- and multi-output cases while having orders of magnitude fewer parameters than competing methods such as neural operators. In fact, we demonstrate that our zero-mean GPs provide competitive results even without training, i.e., we build the first zero-shot mechanism for operator learning. With NN-mean GPs, either in data-driven or physics-informed applications, our strategies eliminate the need for optimizing the kernel parameters and result in a model that augments neural operators with the strengths of kernels. Using FNO and DeepONet as examples, we show that such an integrated model improves the performance of neural operators.

In all the examples in \Cref{sec results} we use the simple kernels in \Cref{eq all kernels} that have at most a few hundred parameters and, hence, limited learning capacity. We studied various kernels including deep ones \cite{RN1901} but did not observe consistent and significant performance improvements. We believe a promising future direction is designing special kernels whose parameters can be efficiently estimated via cross-validation which is more robust to model misspecification compared to MLE. Another interesting direction is leveraging sparse GPs \cite{RN817, RN415, RN413, RN1867, RN414, RN893} that have been developed to increase the scalability of GPs to large datasets. 

\section*{Acknowledgments}
The authors appreciate the support from the Office of Naval Research (grant number $N000142312485$) and the National Science Foundation (grant number $2238038$). HO acknowledges support from a Department of Defense 
Vannevar Bush Faculty Fellowship.

    \printbibliography

@article{micchelli2005learning,
  title={On learning vector-valued functions},
  author={Micchelli, Charles A and Pontil, Massimiliano},
  journal={Neural computation},
  volume={17},
  number={1},
  pages={177--204},
  year={2005},
  publisher={MIT Press}
}

@article{lowery2024kernel,
  title={Kernel Neural Operators (KNOs) for Scalable, Memory-efficient, Geometrically-flexible Operator Learning},
  author={Lowery, Matthew and Turnage, John and Morrow, Zachary and Jakeman, John D and Narayan, Akil and Zhe, Shandian and Shankar, Varun},
  journal={arXiv preprint arXiv:2407.00809},
  year={2024}
}

@article{owhadi2023ideas,
  title={Do ideas have shape? Idea registration as the continuous limit of artificial neural networks},
  author={Owhadi, Houman},
  journal={Physica D: Nonlinear Phenomena},
  volume={444},
  pages={133592},
  year={2023},
  publisher={Elsevier}
}

@article{chen2021solving,
  title={Solving and learning nonlinear PDEs with Gaussian processes},
  author={Chen, Yifan and Hosseini, Bamdad and Owhadi, Houman and Stuart, Andrew M},
  journal={Journal of Computational Physics},
  volume={447},
  pages={110668},
  year={2021},
  publisher={Elsevier}
}

@article{alvarez2012kernels,
  title={Kernels for vector-valued functions: A review},
  author={Alvarez, Mauricio A and Rosasco, Lorenzo and Lawrence, Neil D and others},
  journal={Foundations and Trends{\textregistered} in Machine Learning},
  volume={4},
  number={3},
  pages={195--266},
  year={2012},
  publisher={Now Publishers, Inc.}
}

@article{owhadi2022computational,
  title={Computational graph completion},
  author={Owhadi, Houman},
  journal={Research in the Mathematical Sciences},
  volume={9},
  number={2},
  pages={27},
  year={2022},
  publisher={Springer}
}

@article{breiman1997predicting,
  title={Predicting multivariate responses in multiple linear regression},
  author={Breiman, Leo and Friedman, Jerome H},
  journal={Journal of the Royal Statistical Society Series B: Statistical Methodology},
  volume={59},
  number={1},
  pages={3--54},
  year={1997},
  publisher={Oxford University Press}
}

@inproceedings{evgeniou2004regularized,
  title={Regularized multi--task learning},
  author={Evgeniou, Theodoros and Pontil, Massimiliano},
  booktitle={Proceedings of the tenth ACM SIGKDD international conference on Knowledge discovery and data mining},
  pages={109--117},
  year={2004}
}

@article{argyriou2008convex,
  title={Convex multi-task feature learning},
  author={Argyriou, Andreas and Evgeniou, Theodoros and Pontil, Massimiliano},
  journal={Machine learning},
  volume={73},
  pages={243--272},
  year={2008},
  publisher={Springer}
}

@misc{RN499,
   ISBN = {0486428184},
   year = {2003},
   type = {Generic}
}

@inproceedings{RN2011,
   author = {Hao, Zhongkai and Wang, Zhengyi and Su, Hang and Ying, Chengyang and Dong, Yinpeng and Liu, Songming and Cheng, Ze and Song, Jian and Zhu, Jun},
   title = {Gnot: A general neural operator transformer for operator learning},
   booktitle = {International Conference on Machine Learning},
   publisher = {PMLR},
   pages = {12556-12569},
   ISBN = {2640-3498},
   type = {Conference Proceedings}
}

@article{ramsay1991some,
  title={Some tools for functional data analysis},
  author={Ramsay, James O and Dalzell, CJ1125714},
  journal={Journal of the Royal Statistical Society Series B: Statistical Methodology},
  volume={53},
  number={3},
  pages={539--561},
  year={1991},
  publisher={Oxford University Press}
}

@article{RN1257,
   author = {Li, Zongyi and Kovachki, Nikola and Azizzadenesheli, Kamyar and Liu, Burigede and Bhattacharya, Kaushik and Stuart, Andrew and Anandkumar, Anima},
   title = {Fourier neural operator for parametric partial differential equations},
   journal = {arXiv preprint arXiv:2010.08895},
   year = {2020},
   type = {Journal Article}
}

@article{RN2007,
   author = {Li, Zongyi and Zheng, Hongkai and Kovachki, Nikola and Jin, David and Chen, Haoxuan and Liu, Burigede and Azizzadenesheli, Kamyar and Anandkumar, Anima},
   title = {Physics-informed neural operator for learning partial differential equations},
   journal = {ACM/JMS Journal of Data Science},
   ISSN = {2831-3194},
   year = {2021},
   type = {Journal Article}
}

@article{RN1380,
   author = {Lu, Lu and Jin, Pengzhan and Karniadakis, George Em},
   title = {Deeponet: Learning nonlinear operators for identifying differential equations based on the universal approximation theorem of operators},
   journal = {arXiv preprint arXiv:1910.03193},
   year = {2019},
   type = {Journal Article}
}

@article{RN2058,
   author = {Owhadi, Houman},
   title = {Bayesian numerical homogenization},
   journal = {Multiscale Modeling \& Simulation},
   volume = {13},
   number = {3},
   pages = {812-828},
   ISSN = {1540-3459},
   year = {2015},
   type = {Journal Article}
}

@book{RN2056,
   author = {Owhadi, Houman and Scovel, Clint},
   title = {Operator-Adapted Wavelets, Fast Solvers, and Numerical Homogenization: From a Game Theoretic Approach to Numerical Approximation and Algorithm Design},
   publisher = {Cambridge University Press},
   volume = {35},
   ISBN = {1108484360},
   year = {2019},
   type = {Book}
}

@article{RN2010,
   author = {Serrano, Louis and Le Boudec, Lise and Kassaï Koupaï, Armand and Wang, Thomas X and Yin, Yuan and Vittaut, Jean-Noël and Gallinari, Patrick},
   title = {Operator learning with neural fields: Tackling PDEs on general geometries},
   journal = {Advances in Neural Information Processing Systems},
   volume = {36},
   year = {2024},
   type = {Journal Article}
}

@article{RN563,
   author = {Spanos, P. D. and Ghanem, Roger},
   title = {Stochastic Finite Element Expansion for Random Media},
   journal = {Journal of Engineering Mechanics},
   volume = {115},
   number = {5},
   pages = {1035-1053},
   ISSN = {0733-9399
1943-7889},
   DOI = {10.1061/(asce)0733-9399(1989)115:5(1035)},
   url = {<Go to ISI>://WOS:A1989U235700010},
   year = {1989},
   type = {Journal Article}
}

@article{RN1957,
   author = {Betancourt, José and Bachoc, François and Klein, Thierry and Idier, Déborah and Pedreros, Rodrigo and Rohmer, Jérémy},
   title = {Gaussian process metamodeling of functional-input code for coastal flood hazard assessment},
   journal = {Reliability Engineering \& System Safety},
   volume = {198},
   pages = {106870},
   ISSN = {0951-8320},
   DOI = {https://doi.org/10.1016/j.ress.2020.106870},
   url = {https://www.sciencedirect.com/science/article/pii/S0951832019301693},
   year = {2020},
   type = {Journal Article}
}

@article{RN1275,
   author = {Cutajar, Kurt and Pullin, Mark and Damianou, Andreas and Lawrence, Neil and González, Javier},
   title = {Deep gaussian processes for multi-fidelity modeling},
   journal = {arXiv preprint arXiv:1903.07320},
   year = {2019},
   type = {Journal Article}
}

@article{RN1838,
   author = {Eweis-Labolle, Jonathan Tammer and Oune, Nicholas and Bostanabad, Ramin},
   title = {Data Fusion With Latent Map Gaussian Processes},
   journal = {Journal of Mechanical Design},
   volume = {144},
   number = {9},
   ISSN = {1050-0472},
   DOI = {10.1115/1.4054520},
   url = {https://doi.org/10.1115/1.4054520},
   year = {2022},
   type = {Journal Article}
}

@misc{RN1479,
  title={Kernel interpolation for scalable online Gaussian processes},
  author={Stanton, Samuel and Maddox, Wesley and Delbridge, Ian and Wilson, Andrew Gordon},
  booktitle={International Conference on Artificial Intelligence and Statistics},
  pages={3133--3141},
  year={2021},
  organization={PMLR}
}

@article{RN1845,
   author = {Zanjani Foumani, Zahra and Shishehbor, Mehdi and Yousefpour, Amin and Bostanabad, Ramin},
   title = {Multi-fidelity cost-aware Bayesian optimization},
   journal = {Computer Methods in Applied Mechanics and Engineering},
   volume = {407},
   pages = {115937},
   ISSN = {0045-7825},
   DOI = {https://doi.org/10.1016/j.cma.2023.115937},
   url = {https://www.sciencedirect.com/science/article/pii/S0045782523000609},
   year = {2023},
   type = {Journal Article}
}

@article{RN1883,
   author = {Chen, Yifan and Owhadi, Houman and Schäfer, Florian},
   title = {Sparse Cholesky factorization for solving nonlinear PDEs via Gaussian processes},
   journal = {arXiv preprint arXiv:2304.01294},
   year = {2023},
   type = {Journal Article}
}

@article{RN1890,
   author = {Meng, Rui and Yang, Xianjin},
   title = {Sparse Gaussian processes for solving nonlinear PDEs},
   journal = {Journal of Computational Physics},
   volume = {490},
   pages = {112340},
   ISSN = {0021-9991},
   year = {2023},
   type = {Journal Article}
}

@article{RN1878,
   author = {Yang, Xianjin and Owhadi, Houman},
   title = {A Mini-Batch Method for Solving Nonlinear PDEs with Gaussian Processes},
   journal = {arXiv preprint arXiv:2306.00307},
   year = {2023},
   type = {Journal Article}
}

@article{RN1886,
   author = {Chen, Yifan and Hosseini, Bamdad and Owhadi, Houman and Stuart, Andrew M},
   title = {Solving and learning nonlinear PDEs with Gaussian processes},
   journal = {Journal of Computational Physics},
   volume = {447},
   pages = {110668},
   ISSN = {0021-9991},
   year = {2021},
   type = {Journal Article}
}

@article{RN1919,
   author = {Iwata, Tomoharu and Ghahramani, Zoubin},
   title = {Improving output uncertainty estimation and generalization in deep learning via neural network Gaussian processes},
   journal = {arXiv preprint arXiv:1707.05922},
   year = {2017},
   type = {Journal Article}
}

@article{RN1935,
   author = {Yousefpour, Amin and Foumani, Zahra Zanjani and Shishehbor, Mehdi and Mora, Carlos and Bostanabad, Ramin},
   title = {GP+: A Python Library for Kernel-based learning via Gaussian Processes},
   journal = {arXiv preprint arXiv:2312.07694},
   year = {2023},
   type = {Journal Article}
}

@article{RN1873,
   author = {Zhang, Jiahao and Zhang, Shiqi and Lin, Guang},
   title = {PAGP: A physics-assisted Gaussian process framework with active learning for forward and inverse problems of partial differential equations},
   journal = {arXiv preprint arXiv:2204.02583},
   year = {2022},
   type = {Journal Article}
}

@article{RN1558,
   author = {Planas, R. and Oune, N. and Bostanabad, R.},
   title = {Evolutionary Gaussian Processes},
   journal = {Journal of Mechanical Design},
   volume = {143},
   number = {11},
   pages = {111703},
   ISSN = {1050-0472},
   DOI = {Artn 111703
10.1115/1.4050746},
   url = {<Go to ISI>://WOS:000702466200013},
   year = {2021},
   type = {Journal Article}
}

@book{RN332,
   author = {Rasmussen, Carl Edward},
   title = {Gaussian processes for machine learning},
   year = {2006},
   type = {Book}
}

@inproceedings{RN1901,
   author = {Wilson, Andrew Gordon and Hu, Zhiting and Salakhutdinov, Ruslan and Xing, Eric P},
   title = {Deep kernel learning},
   booktitle = {Artificial intelligence and statistics},
   publisher = {PMLR},
   pages = {370-378},
   type = {Conference Proceedings}
}

@article{RN893,
   author = {Hensman, James and Fusi, Nicolo and Lawrence, Neil D},
   title = {Gaussian processes for big data},
   journal = {arXiv preprint arXiv:1309.6835},
   year = {2013},
   type = {Journal Article}
}

@inproceedings{RN415,
   author = {Seeger, Matthias and Williams, Christopher and Lawrence, Neil},
   title = {Fast forward selection to speed up sparse Gaussian process regression},
   booktitle = {Artificial Intelligence and Statistics 9},
   type = {Conference Proceedings}
}

@inproceedings{RN413,
   author = {Snelson, Edward and Ghahramani, Zoubin},
   title = {Sparse Gaussian processes using pseudo-inputs},
   booktitle = {Advances in neural information processing systems},
   pages = {1257-1264},
   type = {Conference Proceedings}
}

@article{RN414,
   author = {Snelson, Edward and Ghahramani, Zoubin},
   title = {Variable noise and dimensionality reduction for sparse Gaussian processes},
   journal = {arXiv preprint arXiv:1206.6873},
   year = {2012},
   type = {Journal Article}
}

@misc{RN1867,
   volume = {5},
   pages = {567--574},
   url = {https://proceedings.mlr.press/v5/titsias09a.html},
   year = {2009},
   type = {Conference Paper}
}

@article{RN817,
   author = {Williams, Christopher KI and Rasmussen, Carl Edward and Scwaighofer, A and Tresp, Volker},
   title = {Observations on the Nyström method for Gaussian process prediction},
   year = {2002},
   type = {Journal Article}
}

@ARTICLE{Jasak20071,
	author = {Jasak, H. and Tukovic, Z.},
	title = {Automatic mesh motion for the unstructured finite volume method},
	year = {2007},
	journal = {Transactions of FAMENA},
	volume = {30},
	number = {2},
	pages = {1 – 18},
	url = {https://www.scopus.com/inward/record.uri?eid=2-s2.0-45749085537&partnerID=40&md5=cc5c997243b5f9a53b0c3696b6612576},
	type = {Article},
	publication_stage = {Final},
	source = {Scopus},
	note = {Cited by: 2}
}

@article{wang2023discovery,
  title={Discovery of PDEs driven by data with sharp gradient or discontinuity},
  author={Wang, Kang and Zhang, Lei and Tang, Shaoqiang},
  journal={Computers \& Mathematics with Applications},
  volume={140},
  pages={33--43},
  year={2023},
  publisher={Elsevier}
}

@article{RN2059,
   author = {Tripura, Tapas and Chakraborty, Souvik},
   title = {Wavelet Neural Operator for solving parametric partial differential equations in computational mechanics problems},
   journal = {Computer Methods in Applied Mechanics and Engineering},
   volume = {404},
   pages = {115783},
   ISSN = {0045-7825},
   DOI = {https://doi.org/10.1016/j.cma.2022.115783},
   url = {https://www.sciencedirect.com/science/article/pii/S0045782522007393},
   year = {2023},
   type = {Journal Article}
}

@article{RN2054,
   author = {Wang, Sifan and Wang, Hanwen and Perdikaris, Paris},
   title = {Learning the solution operator of parametric partial differential equations with physics-informed DeepONets},
   journal = {Science advances},
   volume = {7},
   number = {40},
   pages = {eabi8605},
   ISSN = {2375-2548},
   year = {2021},
   type = {Journal Article}
}

@book{RN2055,
   author = {Xiu, Dongbin},
   title = {Numerical methods for stochastic computations: a spectral method approach},
   publisher = {Princeton university press},
   ISBN = {1400835348},
   year = {2010},
   type = {Book}
}

@article{RN2057,
   author = {Xiu, Dongbin and Shen, Jie},
   title = {Efficient stochastic Galerkin methods for random diffusion equations},
   journal = {Journal of Computational Physics},
   volume = {228},
   number = {2},
   pages = {266-281},
   ISSN = {0021-9991},
   year = {2009},
   type = {Journal Article}
}

@article{RN1421,
   author = {Chen, T. and Chen, H.},
   title = {Universal approximation to nonlinear operators by neural networks with arbitrary activation functions and its application to dynamical systems},
   journal = {IEEE Trans Neural Netw},
   volume = {6},
   number = {4},
   pages = {911-7},
   ISSN = {1045-9227 (Print)
1045-9227 (Linking)},
   DOI = {10.1109/72.392253},
   url = {https://www.ncbi.nlm.nih.gov/pubmed/18263379},
   year = {1995},
   type = {Journal Article}
}

@article{RN1508,
   author = {Li, Zongyi and Kovachki, Nikola and Azizzadenesheli, Kamyar and Liu, Burigede and Bhattacharya, Kaushik and Stuart, Andrew and Anandkumar, Anima},
   title = {Multipole graph neural operator for parametric partial differential equations},
   journal = {arXiv preprint arXiv:2006.09535},
   year = {2020},
   type = {Journal Article}
}

@article{RN1509,
   author = {Li, Zongyi and Kovachki, Nikola and Azizzadenesheli, Kamyar and Liu, Burigede and Bhattacharya, Kaushik and Stuart, Andrew and Anandkumar, Anima},
   title = {Neural operator: Graph kernel network for partial differential equations},
   journal = {arXiv preprint arXiv:2003.03485},
   year = {2020},
   type = {Journal Article}
}

@article{RN2053,
   author = {He, Juncai and Liu, Xinliang and Xu, Jinchao},
   title = {MgNO: Efficient parameterization of linear operators via multigrid},
   journal = {arXiv preprint arXiv:2310.19809},
   year = {2023},
   type = {Journal Article}
}

@book{RN2060,
   author = {Hackbusch, Wolfgang},
   title = {Multi-grid methods and applications},
   publisher = {Springer Science \& Business Media},
   volume = {4},
   ISBN = {3662024276},
   year = {2013},
   type = {Book}
}

@article{RN2061,
   author = {He, Juncai and Xu, Jinchao},
   title = {MgNet: A unified framework of multigrid and convolutional neural network},
   journal = {Science china mathematics},
   volume = {62},
   pages = {1331-1354},
   ISSN = {1674-7283},
   year = {2019},
   type = {Journal Article}
}

@article{RN287,
   author = {Conti, Stefano and O’Hagan, Anthony},
   title = {Bayesian emulation of complex multi-output and dynamic computer models},
   journal = {Journal of Statistical Planning and Inference},
   volume = {140},
   number = {3},
   pages = {640-651},
   note = {531at
Times Cited:195
Cited References Count:30},
   ISSN = {03783758},
   DOI = {10.1016/j.jspi.2009.08.006},
   url = {<Go to ISI>://WOS:000272635800005},
   year = {2010},
   type = {Journal Article}
}

@article{pgcan2024,
  author = {Shishehbor, Mehdi and Hosseinmardi, Shirin and Bostanabad, Ramin},
  title = {Parametric encoding with attention and convolution mitigate spectral bias of neural partial differential equation solvers},
  journal = {Structural and Multidisciplinary Optimization},
  year = {2024},
  volume = {67},
  number = {7},
  pages = {128},
  issn={1615-1488},
  doi={10.1007/s00158-024-03834-7},
}

@article{ramsay2005fitting,
  title={Fitting differential equations to functional data: Principal differential analysis},
  author={Ramsay, James O and Silverman, Bernard W},
  journal={Functional data analysis},
  pages={327--348},
  year={2005},
  publisher={Springer}
}

@book{ramsay2002applied,
  title={Applied functional data analysis: methods and case studies},
  author={Ramsay, James O and Silverman, Bernard W},
  year={2002},
  publisher={Springer}
}

@book{ghanem2003stochastic,
  title={Stochastic finite elements: a spectral approach},
  author={Ghanem, Roger G and Spanos, Pol D},
  year={2003},
  publisher={Courier Corporation}
}

@article{lucia2004reduced,
  title={Reduced-order modeling: new approaches for computational physics},
  author={Lucia, David J and Beran, Philip S and Silva, Walter A},
  journal={Progress in aerospace sciences},
  volume={40},
  number={1-2},
  pages={51--117},
  year={2004},
  publisher={Elsevier}
}

@inproceedings{mitra2009acoustics,
  title={From acoustics to vocal tract time functions},
  author={Mitra, Vikramjit and Ozbek, Yucel and Nam, Hosung and Zhou, Xinhui and Espy-Wilson, Carol Y},
  booktitle={2009 IEEE International Conference on Acoustics, Speech and Signal Processing},
  pages={4497--4500},
  year={2009},
  organization={IEEE}
}

@article{economon2016su2,
  title={SU2: An open-source suite for multiphysics simulation and design},
  author={Economon, Thomas D and Palacios, Francisco and Copeland, Sean R and Lukaczyk, Trent W and Alonso, Juan J},
  journal={Aiaa Journal},
  volume={54},
  number={3},
  pages={828--846},
  year={2016},
  publisher={American Institute of Aeronautics and Astronautics}
}

@article{kovachki2022multiscale,
  title={Multiscale modeling of materials: Computing, data science, uncertainty and goal-oriented optimization},
  author={Kovachki, Nikola and Liu, Burigede and Sun, Xingsheng and Zhou, Hao and Bhattacharya, Kaushik and Ortiz, Michael and Stuart, Andrew},
  journal={Mechanics of Materials},
  volume={165},
  pages={104156},
  year={2022},
  publisher={Elsevier}
}

@article{boncoraglio2021active,
  title={Active manifold and model-order reduction to accelerate multidisciplinary analysis and optimization},
  author={Boncoraglio, Gabriele and Farhat, Charbel},
  journal={AIAA Journal},
  volume={59},
  number={11},
  pages={4739--4753},
  year={2021},
  publisher={American Institute of Aeronautics and Astronautics}
}

@article{lu2021learning,
  title={Learning nonlinear operators via DeepONet based on the universal approximation theorem of operators},
  author={Lu, Lu and Jin, Pengzhan and Pang, Guofei and Zhang, Zhongqiang and Karniadakis, George Em},
  journal={Nature machine intelligence},
  volume={3},
  number={3},
  pages={218--229},
  year={2021},
  publisher={Nature Publishing Group UK London}
}

@inproceedings{raonic2023convolutional,
  title={Convolutional neural operators},
  author={Raonic, Bogdan and Molinaro, Roberto and Rohner, Tobias and Mishra, Siddhartha and de Bezenac, Emmanuel},
  booktitle={ICLR 2023 Workshop on Physics for Machine Learning},
  year={2023}
}

@article{li2020fourier,
  title={Fourier neural operator for parametric partial differential equations},
  author={Li, Zongyi and Kovachki, Nikola and Azizzadenesheli, Kamyar and Liu, Burigede and Bhattacharya, Kaushik and Stuart, Andrew and Anandkumar, Anima},
  journal={arXiv preprint arXiv:2010.08895},
  year={2020}
}

@article{batlle2024kernel,
  title={Kernel methods are competitive for operator learning},
  author={Batlle, Pau and Darcy, Matthieu and Hosseini, Bamdad and Owhadi, Houman},
  journal={Journal of Computational Physics},
  volume={496},
  pages={112549},
  year={2024},
  publisher={Elsevier}
}

@inproceedings{hao2023gnot,
  title={Gnot: A general neural operator transformer for operator learning},
  author={Hao, Zhongkai and Wang, Zhengyi and Su, Hang and Ying, Chengyang and Dong, Yinpeng and Liu, Songming and Cheng, Ze and Song, Jian and Zhu, Jun},
  booktitle={International Conference on Machine Learning},
  pages={12556--12569},
  year={2023},
  organization={PMLR}
}

@article{mora2024neural,
  title={Neural Networks with Kernel-Weighted Corrective Residuals for Solving Partial Differential Equations},
  author={Mora, Carlos and Yousefpour, Amin and Hosseinmardi, Shirin and Bostanabad, Ramin},
  journal={arXiv preprint arXiv:2401.03492},
  year={2024}
}

@article{wang2021understanding,
  title={Understanding and mitigating gradient flow pathologies in physics-informed neural networks},
  author={Wang, Sifan and Teng, Yujun and Perdikaris, Paris},
  journal={SIAM Journal on Scientific Computing},
  volume={43},
  number={5},
  pages={A3055--A3081},
  year={2021},
  publisher={SIAM}
}

@article{baydin2018automatic,
  title={Automatic differentiation in machine learning: a survey},
  author={Baydin, Atilim Gunes and Pearlmutter, Barak A and Radul, Alexey Andreyevich and Siskind, Jeffrey Mark},
  journal={Journal of Marchine Learning Research},
  volume={18},
  pages={1--43},
  year={2018},
  publisher={Microtome Publishing}
}

@article{de2022cost,
  title={The cost-accuracy trade-off in operator learning with neural networks},
  author={de Hoop, Maarten V and Huang, Daniel Zhengyu and Qian, Elizabeth and Stuart, Andrew M},
  journal={arXiv preprint arXiv:2203.13181},
  year={2022}
}

@article{lu2022comprehensive,
  title={A comprehensive and fair comparison of two neural operators (with practical extensions) based on fair data},
  author={Lu, Lu and Meng, Xuhui and Cai, Shengze and Mao, Zhiping and Goswami, Somdatta and Zhang, Zhongqiang and Karniadakis, George Em},
  journal={Computer Methods in Applied Mechanics and Engineering},
  volume={393},
  pages={114778},
  year={2022},
  publisher={Elsevier}
}

@article{wang2022mosaic,
  title={Mosaic flows: A transferable deep learning framework for solving PDEs on unseen domains},
  author={Wang, Hengjie and Planas, Robert and Chandramowlishwaran, Aparna and Bostanabad, Ramin},
  journal={Computer Methods in Applied Mechanics and Engineering},
  volume={389},
  pages={114424},
  year={2022},
  publisher={Elsevier}
}

@article{bonilla2007multi,
  title={Multi-task Gaussian process prediction},
  author={Bonilla, Edwin V and Chai, Kian and Williams, Christopher},
  journal={Advances in neural information processing systems},
  volume={20},
  year={2007}
}

@article{frazier2018tutorial,
  title={A tutorial on Bayesian optimization},
  author={Frazier, Peter I},
  journal={arXiv preprint arXiv:1807.02811},
  year={2018}
}

@article{iwata2017improving,
  title={Improving output uncertainty estimation and generalization in deep learning via neural network Gaussian processes},
  author={Iwata, Tomoharu and Ghahramani, Zoubin},
  journal={arXiv preprint arXiv:1707.05922},
  year={2017}
}

@article{li2020neural,
  title={Neural operator: Graph kernel network for partial differential equations},
  author={Li, Zongyi and Kovachki, Nikola and Azizzadenesheli, Kamyar and Liu, Burigede and Bhattacharya, Kaushik and Stuart, Andrew and Anandkumar, Anima},
  journal={arXiv preprint arXiv:2003.03485},
  year={2020}
}

@article{li2024physics,
  title={Physics-informed neural operator for learning partial differential equations},
  author={Li, Zongyi and Zheng, Hongkai and Kovachki, Nikola and Jin, David and Chen, Haoxuan and Liu, Burigede and Azizzadenesheli, Kamyar and Anandkumar, Anima},
  journal={ACM/JMS Journal of Data Science},
  volume={1},
  number={3},
  pages={1--27},
  year={2024},
  publisher={ACM New York, NY}
}

@article{wang2021learning,
  title={Learning the solution operator of parametric partial differential equations with physics-informed DeepONets},
  author={Wang, Sifan and Wang, Hanwen and Perdikaris, Paris},
  journal={Science advances},
  volume={7},
  number={40},
  pages={eabi8605},
  year={2021},
  publisher={American Association for the Advancement of Science}
}

@article{kovachki2024operator,
  title={Operator learning: Algorithms and analysis},
  author={Kovachki, Nikola B and Lanthaler, Samuel and Stuart, Andrew M},
  journal={arXiv preprint arXiv:2402.15715},
  year={2024}
}

@article{boulle2023mathematical,
  title={A mathematical guide to operator learning},
  author={Boull{\'e}, Nicolas and Townsend, Alex},
  journal={arXiv preprint arXiv:2312.14688},
  year={2023}
}

@article{kovachki2023neural,
  title={Neural operator: Learning maps between function spaces with applications to pdes},
  author={Kovachki, Nikola and Li, Zongyi and Liu, Burigede and Azizzadenesheli, Kamyar and Bhattacharya, Kaushik and Stuart, Andrew and Anandkumar, Anima},
  journal={Journal of Machine Learning Research},
  volume={24},
  number={89},
  pages={1--97},
  year={2023}
}

\end{document}